\pdfoutput=1

\documentclass[11pt]{article}

\usepackage[final]{acl}

\usepackage{times}
\usepackage{latexsym}
\usepackage{color}
\usepackage{graphicx}
\usepackage{multirow}
\usepackage{makecell}
\usepackage{subfigure}
\usepackage{algorithm}
\usepackage{algpseudocode}
\usepackage{booktabs} 
\usepackage{amssymb}
\usepackage{amsmath}
\usepackage{bbm}

\usepackage[T1]{fontenc}

\usepackage[utf8]{inputenc}

\usepackage{microtype}

\usepackage{inconsolata}

\title{Privacy Evaluation Benchmarks for NLP Models}

\author{
 \textbf{Wei Huang$^1$},
 \textbf{Yinggui Wang}$^1$\thanks{Corresponding author (wyinggui@gmail.com).},
 \textbf{CenChen$^2$}
\\
 $^1$Ant Group, China \\
 $^2$East China Normal University, China
\\
   hw378176@antgroup.com, wyinggui@gmail.com, cenchen@dase.ecnu.edu.cn
}

\begin{document}
\maketitle
\begin{abstract}
By inducing privacy attacks on NLP models, attackers can obtain sensitive information such as training data and model parameters, etc. Although researchers have studied, in-depth, several kinds of attacks in NLP models, they are non-systematic analyses. It lacks a comprehensive understanding of the impact caused by the attacks. For example, we must consider which scenarios can apply to which attacks, what the common factors are that affect the performance of different attacks, the nature of the relationships between different attacks, and the influence of various datasets and models on the effectiveness of the attacks, etc. Therefore, we need a benchmark to holistically assess the privacy risks faced by NLP models. In this paper, we present a privacy attack and defense evaluation benchmark in the field of NLP, which includes the conventional/small models and large language models (LLMs). This benchmark supports a variety of models, datasets, and protocols, along with standardized modules for comprehensive evaluation of attacks and defense strategies. Based on the above framework, we present a study on the association between auxiliary data from different domains and the strength of privacy attacks. And we provide an improved attack method in this scenario with the help of Knowledge Distillation (KD).  Furthermore, we propose a chained framework for privacy attacks. Allowing a practitioner to chain multiple attacks to achieve a higher-level attack objective. Based on this, we provide some defense and enhanced attack strategies. 
The code for reproducing the results can be found at https://github.com/user2311717757/nlp\_doctor.
\end{abstract}

\section{Introduction}
In the past few decades, research in the field of NLP-based machine learning, especially deep learning, has achieved significant progress. However, the advancement of these applications has also led to serious security and privacy risks. In particular, inference attacks ~\cite{zhang2020secret,mehnaz2022your,he2022extracted,mireshghallah2022empirical} enable attackers to deduce critical user privacy information such as training data and target model parameters. In general, current privacy attacks are studied under various threat models and experimental setups, but they are typically examined in isolation. It necessitates a comprehensive understanding of the risks these attacks pose, including the common factors influencing their performance, the scenarios where different inference attacks can be applied, the effectiveness of defenses, and the relationships between the attacks. Therefore, we need a benchmark to holistically assess the privacy risks faced by NLP models. To fill this gap, we conduct a comprehensive privacy risk assessment of NLP models targeting four representative inference attacks and have open-sourced an NLP privacy evaluation benchmark.

\begin{figure*}[ht]
    \begin{center}
        \includegraphics[width=0.85\linewidth]{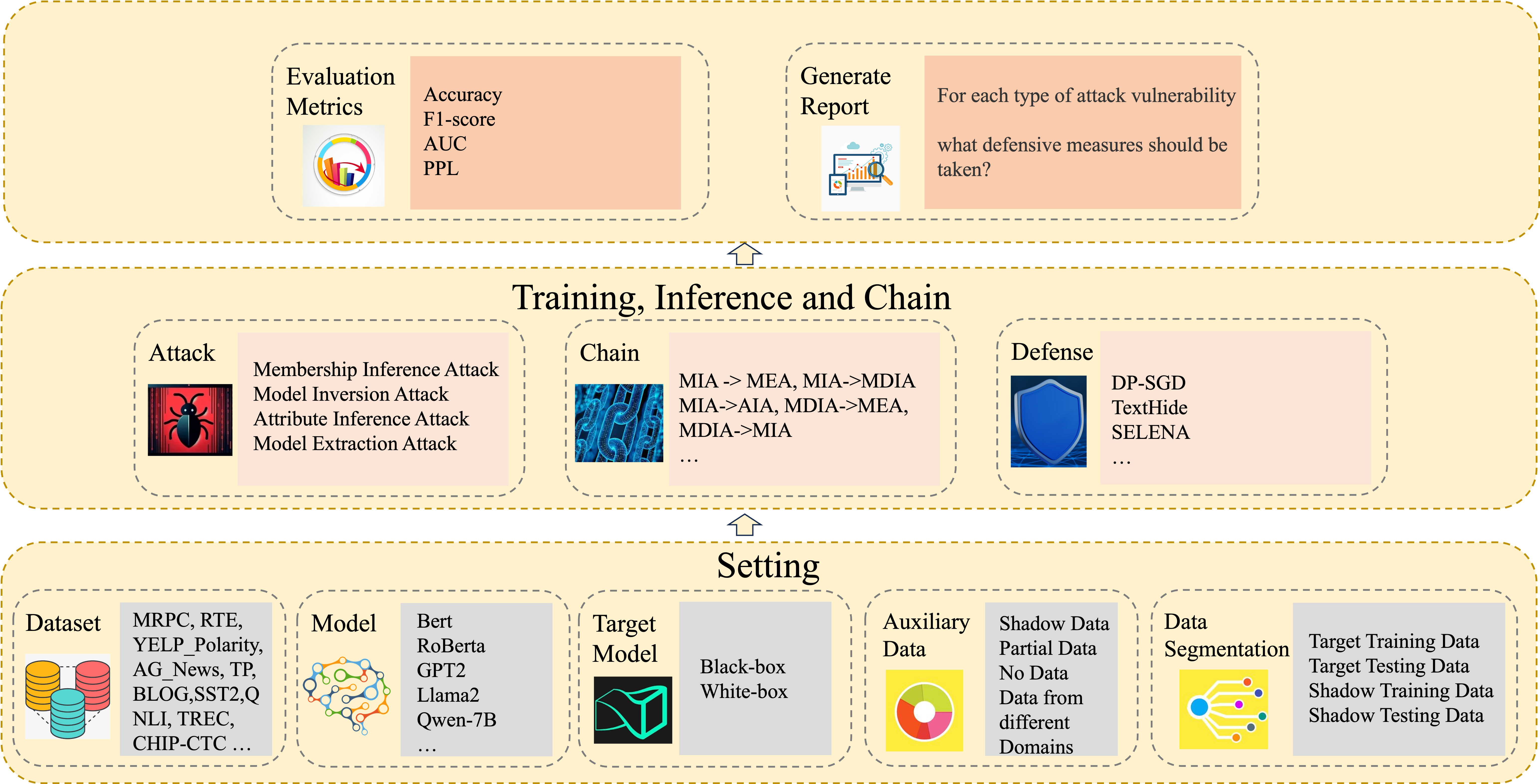}
    \end{center}
    \vspace{-0.3cm}
    \caption{\label{fig:nlp_ad}Overview of our privacy evaluation benchmark for NLP models.}
    \vspace{-1.2em}
    \end{figure*}

\textbf{Attacks and Defenses for NLP models}: 
In this paper, we focus on studying four representative privacy attacks on both large (Llama2~\cite{touvron2023llama}, Qwen~\cite{bai2023qwen}, and GPT2-xl~\cite{radford2019language}) and small (BERT~\cite{devlin2018bert}, RoBERTa~\cite{liu2019roberta}, and GPT2-small~\cite{radford2019language}) NLP models: Membership Inference Attack (MIA, LMIA)~\cite{Shokri2016MembershipIA}, Model Inversion Attack (MDIA, LMDIA)~\cite{zhang2020secret}, Attribute Inference Attack (AIA, LAIA)~\cite{mehnaz2022your}, and Model Extraction Attack (MEA, LMEA)~\cite{he2022extracted}. Adding an "L" before the name indicates an attack targeting LLM.
Besides the four attack methods, we integrate some comprehensive defense methods, including DP-SGD ~\cite{abadi2016deep}, SELENA ~\cite{280000}, and Texthide~\cite{huang2020texthide}. 
Although there is an existing evaluation system for privacy attacks in the image domain~\cite{277098}, our work differs from it in the following ways: 1) This paper focuses on NLP models, not the image-domain ones. 2) We introduce a privacy attack module for LLMs. 3) We explore the impact of auxiliary data from different domains on attacks. 4) We propose a chained framework for privacy attacks.

\textbf{Attacks using Auxiliary Data from Different Domains}: To fit the real scenarios (attackers may not obtain data with the same distribution as the auxiliary data), we conduct extensive experiments with data from different domains.
From the experimental results (see Appendix~\ref{16}), we find that there are some cases where the attack performance is very low.
Through our analysis, we conclude that the main reason for the poor performance of the attack is that the data distribution from different domains may be far from the original data distribution. We use KD to mitigate this issue.

\textbf{ A Chained Framework for Different Attacks}: From an attack perspective, we hope to enable attackers to chain multiple attacks to achieve a higher-level attack objective. Secondly, we aim to explore whether one attack can influence the capability of another, in order to clarify the sequential relationship among different attacks, the inherent correlational characteristics among them, and the possibility of linking different attacks together. Based on the considerations mentioned above, we propose a chained framework for different attacks.
From this, we provide some defense and enhanced attack methods: a defense method against MIA, a data-free MIA, and some strategies to improve the success rate of AIA, MDIA, and MEA.

The main contributions of this paper are as follows:
In order to promote development and practical deployment of NLP privacy evaluation, we present a privacy evaluation benchmark for NLP models, which includes the conventional NLP models and LLMs. This paper experimentally analyzes the relationship between the data from different domains and attack performance. We use KD to address the problem of the low success rate of MIA under different domains, and also propose a chained framework for privacy attacks. It is mainly to enable a practitioner to chain multiple attacks to achieve a higher-level objective or explore the inherent correlational characteristics among attacks. 

\section{Overview of Privacy Evaluation Benchmarks}
\subsection{Implemented Components}
Figure ~\ref{fig:nlp_ad} provides an overview of the privacy evaluation benchmark. It primarily integrates modules for privacy attacks, defenses, and evaluations for various conventional NLP models and LLMs. The benchmark encompasses a wide range of threat models and usage scenarios, covering settings for target models in both black-box and white-box contexts. Additionally, it provides auxiliary data settings with shadow data, partial data, no data, and data from different domains. The benchmark also features data partitioning strategies for target training data, target testing data, shadow training data, and shadow testing data. In addition, we have integrated a chained framework that serves as an extension module for privacy attacks, aiming to explore higher attack objectives and the interrelationships between different attacks. Currently, our privacy evaluation benchmark supports 3 conventional models and 3 LLMs, corresponding to 16 and 9 types of privacy attacks, respectively. In addition, it supports 14 types of chained connections and 3 defense mechanisms, as well as a wealth of evaluation strategies. Altogether, it supports 15 datasets from various tasks, including but not limited to classification and generation. We have provided a comprehensive description of the benchmark's usage process, with intricate details presented in Appendix ~\ref{htub}. Below, we will provide a detailed description of the settings for different threat models, the objectives of each attack, and how to use this benchmark.

\subsection{Model Description}
In this work, we focus on the classification (conventional models) and generation tasks (LLMs) in the NLP domain, which are among the most popular NLP applications. Typically, the goal of NLP classification tasks is to map data samples to a category. 
The output of an NLP classification model is a probability vector. 
And the output of an NLP generation model is text generated by the model based on the language patterns it has learned.
    
\subsection{Threat Model}
In different studies of privacy attacks, researchers have different restrictions on the knowledge that can be accessed by attackers, focusing on two aspects: access to the target model and auxiliary data.

\textbf{Target Model}: Referring to existing work, we can categorize the way an attacker accesses the model into two different setups: black box and white box. The former is denoted as $B^{box}$, which means that the attacker knows nothing about the internal structure of the target model. 
The latter is denoted as $W^{box}$, which means that the attacker has all the information about the target model, including model parameters, structure, and so on. 

\textbf{Auxiliary Data}: 
In this paper, we classify the auxiliary data that an attacker can obtain into four cases: shadow data, partial data, no data, and auxiliary data from
different domains. For the first one, it is denoted as $D^{sha}$, which means that the attacker has the same distributed data as the auxiliary data. 
The second one is written as $D^{par}$, which symbolizes that the adversary can get a part of the target data. For the third one,  $D^{no}$, means that the attacker does not have any auxiliary data. For the fourth one, it is denoted as $D^{diff}$, which means that the attacker can only obtain data that has a different distribution from the training data.

\textbf{Data Segmentation}: To comply with the above assumption for the auxiliary data, we randomly divide the training data set into four non-overlapping sub-datasets of the same size as follows: Target Training Data, Target Testing Data, Shadow Training Data, and Shadow Testing Data. The first one is used to train the target model. The second one is used to evaluate the performance of attacks. The third is used to train the shadow model for attacks and is also used as a query dataset for various attacks. The last one is used in the training of a classification model for MIAs.

\subsection{The Attack Objectives}
Here we formally define the attack objectives under different attacks. The goal of a Membership Inference Attack is for the attacker to determine whether a target data sample was used to train the NLP model. The goal of a Model Inversion Attack is for the attacker to infer the training data of the target model when given access to the target model and auxiliary data, allowing an attacker to directly learn information about the training data. The objective of an Attribute Inference Attack is for the attacker to learn additional attribute information about the training data that is unrelated to the original task, such as gender, age, etc. This attack is used to explore unintentional information leakage. The goal of a Model Extraction Attack is to extract the parameters of the target model. Ideally, an attacker would have the capability to obtain a model with performance very similar to that of the target model, thereby achieving the effect of model reuse. As mentioned above, privacy attacks pose serious privacy threats; therefore, there is an urgent need for a benchmark to assess the privacy risks faced by the NLP model.

\section{Privacy Attacks and Defenses for NLP Models} 
\label{se4}
\subsection{Privacy Attacks for NLP Models}
This paper will focus on four privacy attacks against NLP small models: Membership Inference Attacks (MIA), Model Inversion Attacks (MDIA), Attribute Inference Attacks (AIA), and Model Extraction Attacks (MEA). Due to space limitations, we briefly describe the integrated attacks under various threat models in the main text, with detailed attack specifics provided in the appendices~\ref{1}. Attack methods are summarized in Table ~\ref{ttt}. 

We have integrated eight types of MIAs for different threat models, five of which are black-box attacks ($B^{box}$/($D^{sha}$, $D^{par}$ or $D^{diff}$)~\cite{shejwalkar2021membership,47e5,song2021systematic}), which rely on thresholds or classification models to achieve the attack objectives. There are three white-box attacks (White Box/($D^{sha}$, $D^{par}$ or $D^{diff}$)~\cite{sp/NasrSH19}), which mainly utilize gradients to distinguish between members and non-members. Detailed descriptions are included in the Appendix~\ref{2} and~\ref{6}.

We have integrated three types of MDIAs focusing on white-box settings ($W^{box}$/($D^{no}$~\cite{Fredrikson2015ModelIA}, $D^{sha}$ or $D^{diff}$)~\cite{zhang2022text,iclr/DathathriMLHFMY20}). In our research, we have not yet discovered black-box MDIA on NLP classification models. One can imagine how difficult it would be for an attacker to infer the training data if they can only obtain logits or even just labels. Detailed descriptions are included in the Appendix~\ref{3} and~\ref{6}.

For AIAs, we have implemented two types of attacks based on different threat models: black-box and white-box ($B^{box}$ and $W^{box}$/$D^{sha}$~\cite{Lyu2021KillingTB,iclr/SongS20}). Both attacks rely on training attack models to ascertain attribute information. 
Detailed descriptions are included in the Appendix~\ref{4}.

Our benchmark also includes three types of black-box MEAs ($B^{box}$/($D^{sha}$, $D^{par}$ or $D^{diff}$)~\cite{Lyu2021KillingTB,he2021model}). Since the primary target of such attacks is the model parameters, assuming a white-box target model would negate the purpose of the attack. We primarily focus on models released as APIs. Detailed descriptions are included in the Appendix~\ref{5} and~\ref{6}.

\subsection{Solutions to Low Success Rate of MIA}
From the experimental results (see Appendix ~\ref{16}), we observe that in some cases membership inference performance is 0.500. 
We think that it is mainly because the shadow models trained on data from different domains cannot effectively summarize the membership states of the target model. We thus use KD~\cite{hinton2015distilling} to alleviate the above problem.

\begin{figure}[t]
\begin{center}
    \includegraphics[width=0.9\linewidth]{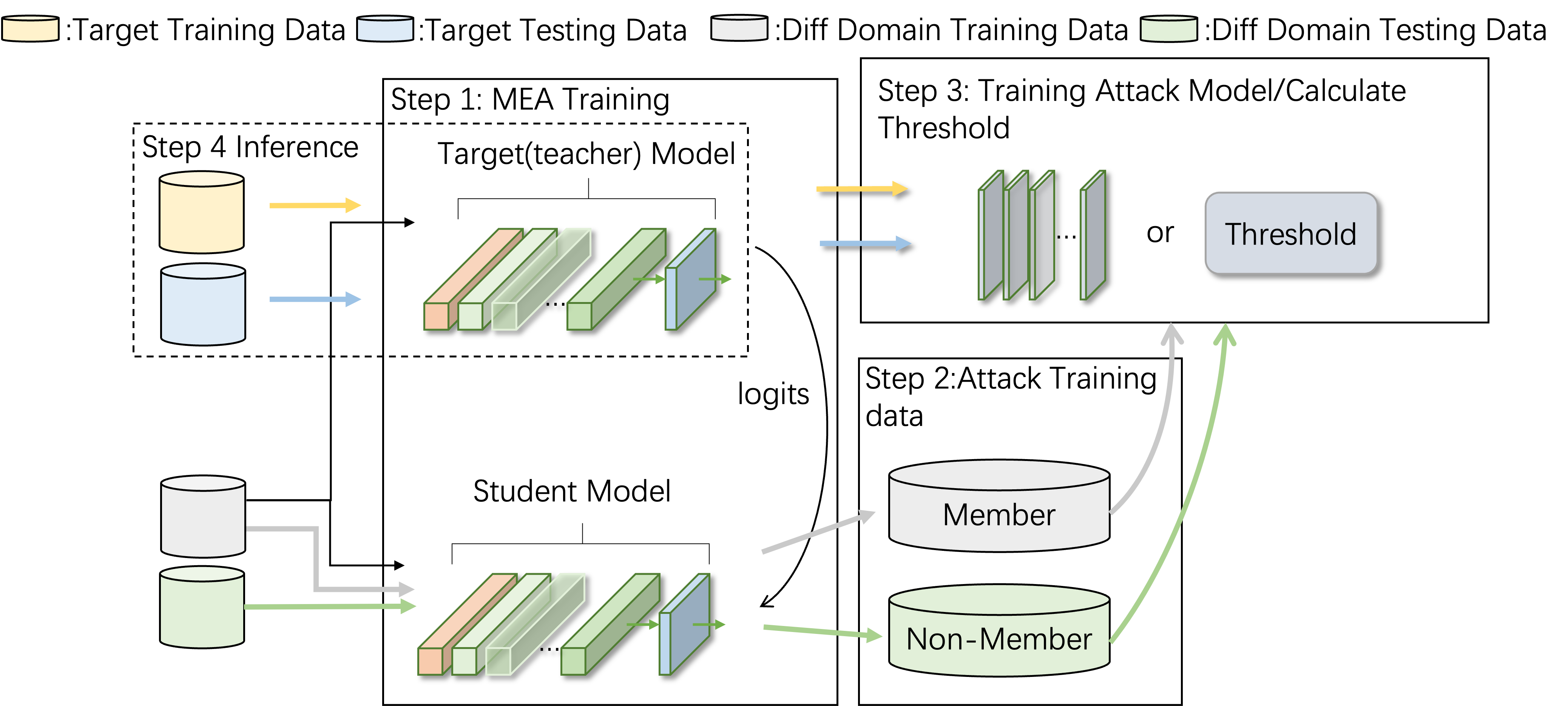}
\end{center}
\vspace{-0.2cm}
\caption{\label{fig:figure1}The flow chart for mitigating the low-performance issue of MIA under data from different domains with the help of KD.}
\vspace{-0.5em}
\end{figure}

We know that KD can distill the knowledge contained in the teacher model into the student model. In other words, the student model can simulate the behavior of the teacher (target) model. 
The specific flow of this strategy is shown in Figure~\ref{fig:figure1}. Concretely, the target model is queried using training data from different domains to obtain logits, followed by training the student model using the data and labels mentioned above. Afterwards, the model is used as a shadow model for MIAs, and finally, the training and inference are performed according to the black-box MIA proposed in Appendix~\ref{2}. 
\subsection{Privacy Defenses for NLP Models}
From the previous introduction, we have learned that privacy attacks can cause serious damage, so defenses against them are essential. Here, we select three classical or SOTA methods, namely, DP-SGD~\cite{abadi2016deep}, SELENA~\cite{280000}, and TextHide~\cite{huang2020texthide}. Detailed descriptions are included in the Appendices~\ref{7}.
\section{The Chained Framework}
\label{sec6}
\begin{figure}[!t]
\begin{center}
    \includegraphics[width=0.8\linewidth]{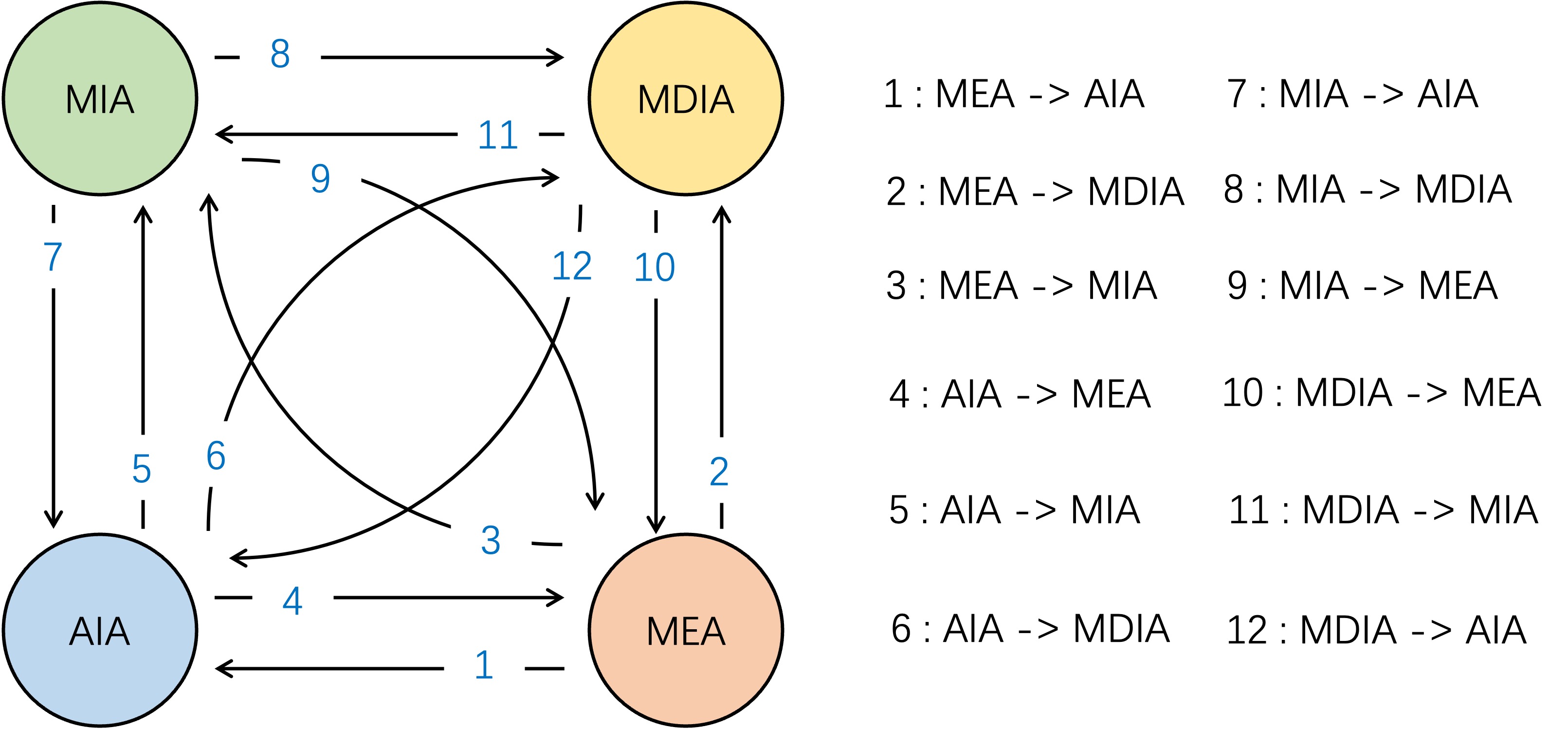}
\end{center}
\vspace{-0.2cm}
\caption{\label{fig:figure2}Framework for chaining different attacks.}
\vspace{-1.2em}
\end{figure}

Most work on privacy attacks focused on a particular type of attack. Whether combining different attacks may yield different results is still an open problem. We propose a chained framework to enable a practitioner to chain multiple attacks to achieve a higher-level objective or explore the inherent correlational characteristics between attacks. Figure ~\ref{fig:figure2} shows the proposed chained framework.

\begin{figure}[!t]
\begin{center}
    \includegraphics[width=1\linewidth]{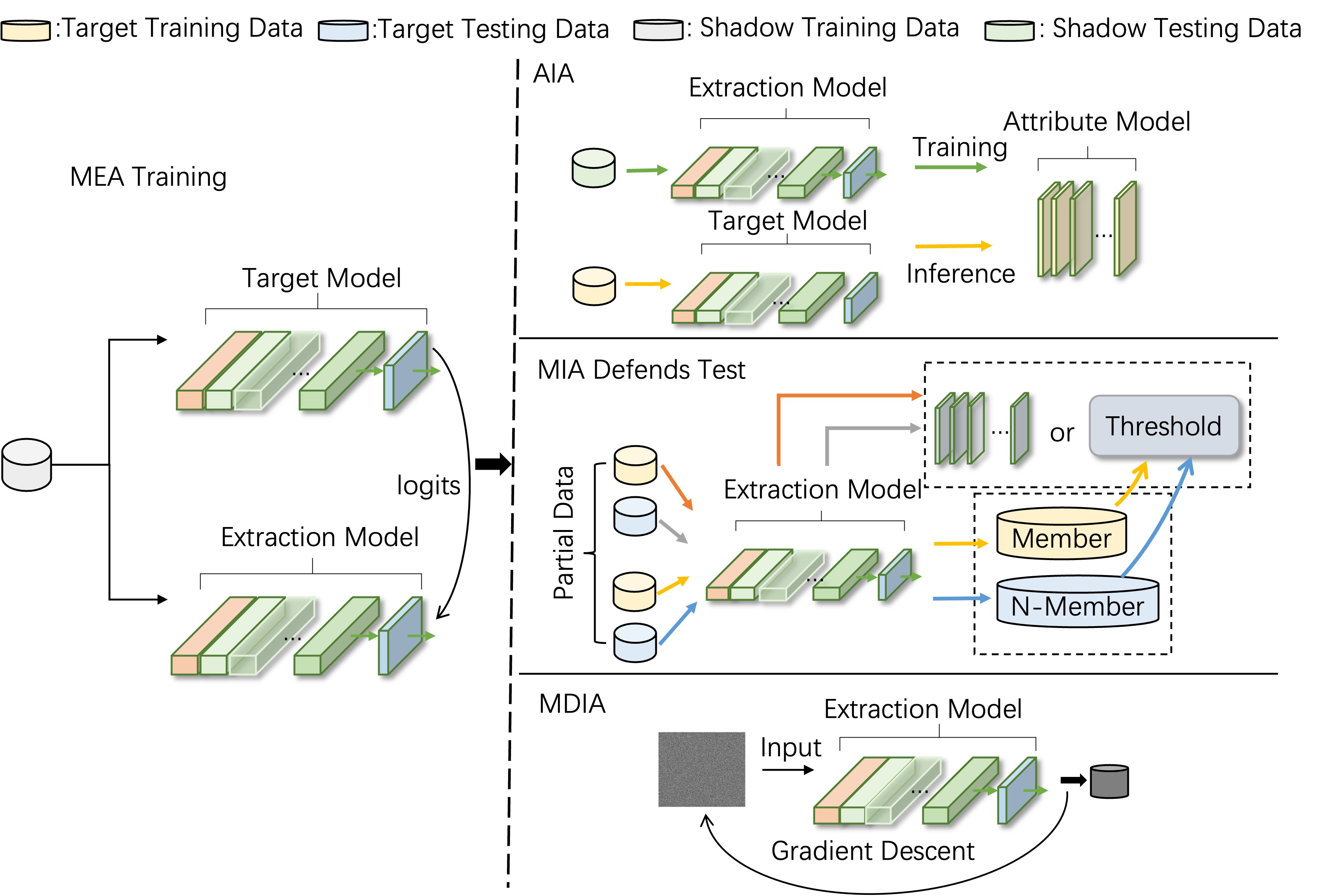}
\end{center}
\vspace{-0.2cm}
\caption{\label{fig:figure22}Flow chart of MEA chained with AIA/MIA/MDIA.}
\vspace{-0.2cm}
\end{figure}

\textbf{MEA chained with AIA/MIA/MDIA}: 
Figure~\ref{fig:figure22} illustrates the flowchart of the method for chaining Model Extraction Attack (MEA) with other attacks. The initial step in this integration process involves utilizing shadow training data to query the target model, to train the extraction model. The subsequent steps are as follows.

\textbf{MEA $\rightarrow$ AIA:} The attacker uses shadow testing data to query the extraction model, which is then employed to train the attribute attack model. Finally, the attacker utilizes the target training data for inference. Since the attacker can access all parameters of the extracted model, they can deploy a white-box AIA method based on this model.

\textbf{MEA $\rightarrow$ MIA:} After the model owner publishes the extracted model, the attacker can proceed with the standard MIA procedure to conduct the attack test. Since MEAs leverage $D^{sha}$ (shadow data) to obtain the extraction model, and $D^{sha}$ does not belong to the target data, the extraction model can simulate the performance of the target model without inheriting the target model's training data. So, we consider that MEA can act as a defense strategy against MIA. 

\textbf{MEA $\rightarrow$ MDIA:} This method employs the extracted model as a pseudo-target model for MDIAs to generate raw data. The purpose is to assess whether the extracted model retains any memory of the original training data.
\begin{figure}[t]
\begin{center}
    \includegraphics[width=1\linewidth]{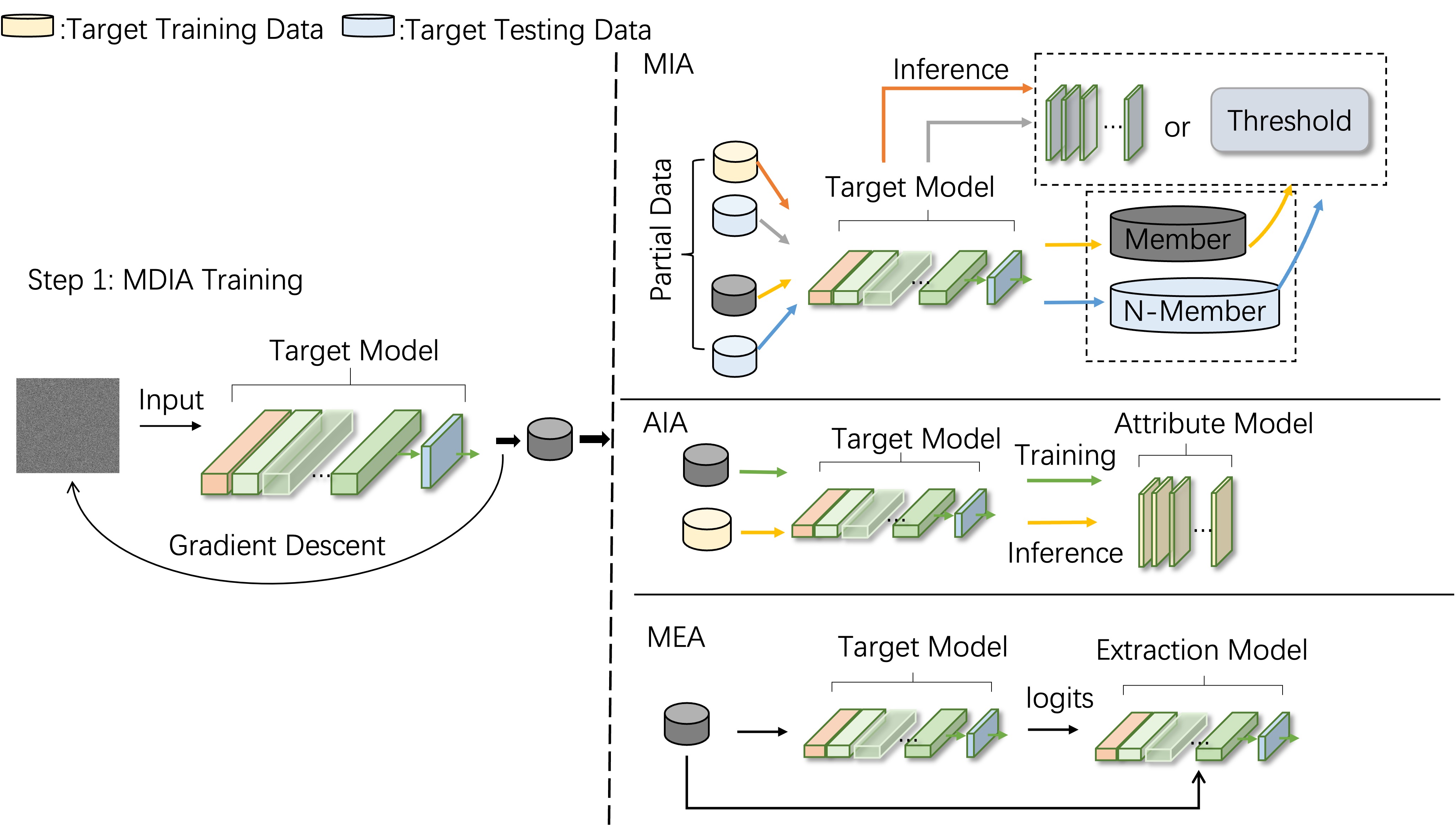}
\end{center}
\vspace{-0.2cm}
\caption{\label{fig:figure3}Flow chart of MDIA chained with MIA/AIA/MEA.}
\vspace{-1.2em}
\end{figure}
\begin{figure}[t]
\begin{center}
    \includegraphics[width=1\linewidth]{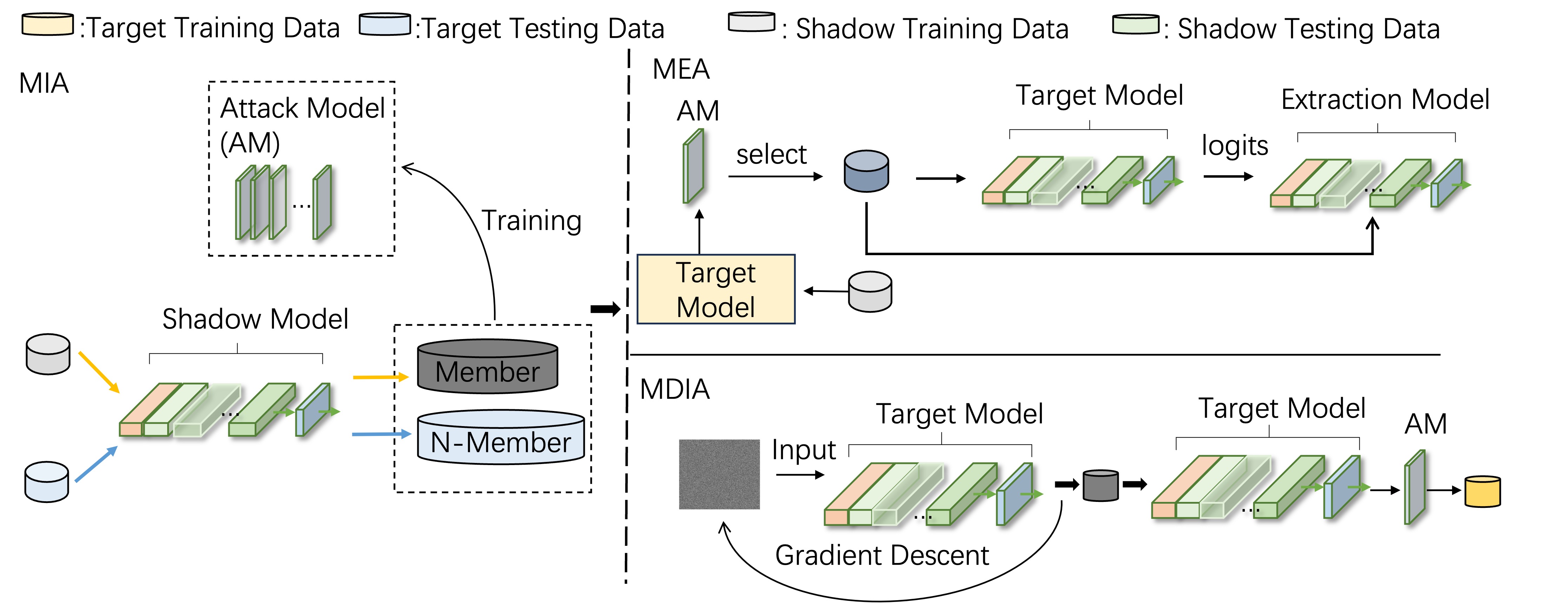}
\end{center}
\vspace{-0.2cm}
\caption{\label{fig:figure33}Flow chart of MIA chained with MEA/MDIA.}
\vspace{-1.2em}
\end{figure} 

\textbf{MDIA chained with MIA/AIA/MEA}: 
Figure~\ref{fig:figure3} depicts the integration of MDIA with the remaining attacks. Given that the MDIA is a data-free reconstruction method, it is anticipated that the derived method can serve as a data generation strategy. This generated data can effectively supplant the assumption that the attacker possesses auxiliary data, thereby facilitating data-free attacks. The steps involved are \textbf{MDIA $\rightarrow$ MIA/AIA/MEA}, i.e., the attacker utilizes the data generated by the MDIA as auxiliary data to conduct the MIA, AIA, or MEA.

\textbf{MIA chained with MEA/MDIA}: 
Figure~\ref{fig:figure33} illustrates chaining MIA with MEA or MDIA. MIA can identify whether a data point is part of the training set, thus it can serve as a data filter. The initial step involves training both the shadow model and the attack model. The subsequent steps include
MIA $\rightarrow$ MEA and MIA $\rightarrow$ MDIA. In the \textbf{MIA $\rightarrow$ MEA} step, the shadow data is first processed by MIA to obtain the logits from the attack model, then the logits' member dimensions are ranked by score magnitude, and finally, the data is selected based on these scores proportionally.
In \textbf{MIA $\rightarrow$ MDIA}, the attacker generates data using the MDIA, after which the generated data are subjected to the MIA testing process. The generated text is accepted only if the attack model classifies it as a member of the training data set.

Note that the attribute is just an implicit characteristic of a sample. It is irrelevant to the original task when training the target model, and it is just unconsciously remembered by the model. Obviously, in the NLP domain, an attacker who has determined a text attribute could do little to affect the performance of the other attacks. 

\section{Privack Attacks for LLM}
\label{sec7}
In this study, we focus on four LLM privacy attacks, i.e., LMIA, LMDIA, LAIA, and LMEA. Detailed attack specifics are provided in the appendix~\ref{8}. Attack methods are summarized in Table ~\ref{ttt}.

\begin{figure*}[!t]
\begin{center}
    \includegraphics[width=1.0\linewidth]{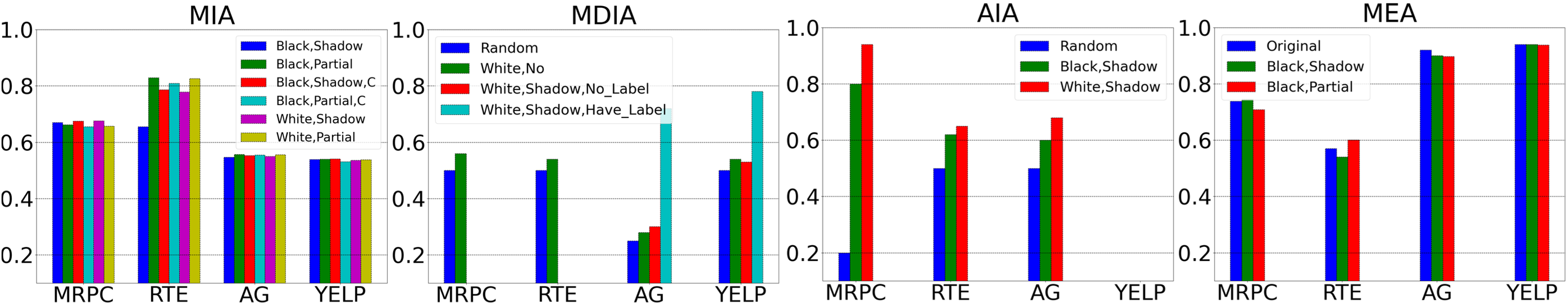}
\end{center}
\setlength{\abovecaptionskip}{-0.1em}
\caption{\label{fig:mia_main}Attack accuracy of MIA/MDIA/AIA/MEA under different data for Bert.}
\vspace{-0.3em}
\end{figure*}

\textbf{ Membership Inference Attack for LLM (LMIA):} Based on different threat assumptions, we introduce three types of LMIAs. The first and second types require shadow data to train a reference model ($B^{box}$/$D^{sha}$)~\cite{mattern2023membership, mireshghallah2022empirical}. The third type ($B^{box}$/$D^{diff}$)~\cite{fu2023practical} considers more realistic scenarios where the attacker can only obtain data from different domains. Detailed descriptions are included in Appendix~\ref{9}.

\textbf{ Model Inversion Attack for LLM (LMDIA):} We introduce two approaches for the black-box LMDIA($D^{no}$ or $D^{diff}$)~\cite{274574}. This paper proposes a language-based LMDIA that employs text generation and Perplexity (PPL) sorting to extract batches of training text. Detailed descriptions are included in Appendix~\ref{10}.

\textbf{Attribute Inference Attack for LLM (LAIA):} We integrated two types of LAIA $B^{box}$/($D^{no}$ or $D^{par}$)~\cite{staab2023beyond,lukas2023analyzing}. This enables the LLM to extract attributes from the texts, including gender, address, occupation, and so on. For more detailed results, refer to Appendix~\ref{11}.

\textbf{ Model Extraction Attack for LLM(LMEA):} This paper employs two attacks suitable for black-box scenarios ($D^{sha}$ or $D^{par}$) ~\cite{tang2023does,gu2023knowledge}. It mainly involves using an LLM to annotate unlabeled data. Detailed descriptions are included in Appendix~\ref{12}.

\section{Results of Attacks and Defenses}

In this section, we describe experimental data, the model, and the results of experiments on NLP privacy attacks and defenses. 

Due to space constraints, we have presented only the most core experimental results in the main text, including privacy attack outcomes for both large-scale and traditional models as well as results from the chained framework. We have placed the performance of using data from different domains (see appendix ~\ref{16}), the experimental results of KD (see appendix ~\ref{erim}), and the defense structures (see appendix ~\ref{17}) in the appendix.

\subsection{Datasets, Models, and Settings}
In this paper, we select fifteen experimental datasets.  
They are: MRPC~\cite{dolan2005automatically}, RTE\cite{wang2019glue}, YELP\_Polarity~\cite{Zhang2015CharacterlevelCN}, AG\_News~\cite{Zhang2015CharacterlevelCN}, TP~\cite{hovy2015user,coavoux2018privacy}, BLOG~\cite{Schler2006EffectsOA,coavoux2018privacy}, SST2~\cite{socher-etal-2013-recursive}, QNLI~\cite{wang2019glue}, TREC~\cite{li2002learning}, CHIP-CTC~\cite{zhang2021cblue}, KUAKE-QIC~\cite{zhang2021cblue}, Wikitext-103 (Wiki)~\cite{merity2016pointer}, ECHR~\cite{chalkidis2019neural}, Enron~\cite{klimt2004introducing} and PersonalReddit (PR)~\cite{staab2023beyond}. 
The specifics of the data can be found in Appendix~\ref{14}.

To verify the effectiveness of different attack methods under different models, six commonly used NLP models, namely BERT~\cite{devlin2018bert}, RoBERTa~\cite{liu2019roberta}, GPT2-small (subsequently abbreviated as GPT2)~\cite{radford2019language}, Qwen-7B~\cite{bai2023qwen}, Llama2 ~\cite{touvron2023llama}and GPT2-xl~\cite{radford2019language} are chosen for this study. We place the experimental setups and performance metrics in Appendix~\ref{14}.

\begin{table*}[ht]
    \centering
    \small
    \scalebox{0.80}{\begin{tabular}{c|c|ccccccccccccc}
    \toprule
        \multicolumn{2}{c}{} & \multirow{2}{*}{Threat model} & \multicolumn{2}{c}{MRPC} & \multicolumn{2}{c}{RTE} & \multicolumn{2}{c}{AG\_News}  & \multicolumn{2}{c}{YELP} & \multicolumn{2}{c}{BLOG} & \multicolumn{2}{c}{TP}\\ \cmidrule{4-15}
        \multicolumn{3}{c}{} & Ori & Acc & Ori & Acc & Ori & Acc & Ori & Acc & Ori & Acc & Ori & Acc\\ \midrule
        \multirow{5}{*}{MEA} & \multirow{2}{*}{AIA} & $<$Black,Shadow$>$ & - & - & - & - & 0.799 & 0.79 & - & - & 0.632 & 0.639 & 0.605 & 0.617 \\
        ~ & ~ &$<$White,Shadow$>$ & - & - & - & - & 0.947 & 0.952 & - & - & 0.659 & 0.647 & 0.681 & 0.663 \\ 
        ~ & MDIA & $<$Black,Shadow$>$ & 0.56 & 0.52 & 0.54 & 0.50 & 0.28 & 0.24 & 0.54 & 0.50 & - & - & - & - \\ 
        ~ & \multirow{2}{*}{MIA} & $<$Black,Partial$>$ & 0.662 & 0.508 & 0.829 & 0.563 & 0.557 & 0.510 & 0.540 & 0.506 & - & - & - & - \\
        ~ & ~ & $<$White,Partial$>$ & 0.657 & 0.506 & 0.826 & 0.547 & 0.556 & 0.515 & 0.538 & 0.504 & - & - & - & - \\ \midrule
        \multirow{5}{*}{MDIA} & \multirow{2}{*}{AIA} & $<$Black,No$>$ & - & - & - & - & 0.799 & 0.540 & - & - & 0.632 & 0.549 & 0.605 & 0.517 \\
        ~ & ~ &$<$White,No$>$ & - & - & - & - & 0.947 & 0.588 & - & - & 0.659 & 0.544 & 0.681 & 0.573 \\ 
        ~ & MEA & $<$Black,No$>$ & - & - & - & - & 0.900 & 0.658 & 0.950 & 0.665 & - & - & - & - \\ 
        ~ & \multirow{2}{*}{MIA} & $<$Black,No$>$ & 0.662 & 0.597 & 0.829 & 0.652 & 0.557 & 0.559 & 0.540 & 0.540 & - & - & - & - \\
        ~ & ~ & $<$White,No$>$ & 0.657 & 0.593 & 0.714 & 0.714 & 0.556 & 0.540 & 0.538 & 0.545 & - & - & - & - \\ \hline
        \multirow{4}{*}{MIA} & \multirow{2}{*}{MEA} & $<$Black,Shadow$>$ & 0.748 & 0.755 & 0.542 & 0.523 & 0.900 & 0.883 & 0.950 & 0.944 & - & - & - & -\\
        ~ & ~ &$<$Black,Partial$>$ & 0.711 & 0.714 & 0.603 & 0.625 & 0.898 & 0.860 & 0.947 & 0.931 & - & - & - & -\\ 
        ~ & \multirow{2}{*}{MDIA} & $<$White,Shadow$>$ & 0.56 & 0.58 & 0.54 & 0.62 & 0.28 & 0.28 & 0.54 & 0.56 & - & - & - & - \\
        ~ & ~ & $<$White,Shadow$>$ & - & - & - & - & 0.30 & 0.30 & 0.53 & 0.54 & - & - & - & -\\ \bottomrule       
    \end{tabular}}
    \caption{Experimental results of the chained framework for the BERT model.}
    \label{tb7}
    \vspace{-0.2em}
\end{table*}

\begin{table*}[!t]
    \centering
    \small
    \scalebox{0.80}{\begin{tabular}{c|cccccc}
    \toprule
     \multirow{2}{*}{Methods} & \multicolumn{2}{c}{KUAKE-IR} & \multicolumn{2}{c}{CHIP-CTC} & \multicolumn{2}{c}{KUAKE-QTR}\\ \cmidrule{2-7}
     ~ &Llama2 &Qwen &Llama2 &Qwen &Llama2 &Qwen \\ 
     \midrule
    LiRA(Black Box/$D^{sha}$) &0.541  &0.535  &0.524     &0.519 &0.557 &0.542  \\ 
    LOSS Attack(Black Box/$D^{sha}$) &0.607  &0.624  &0.572  &0.558 &0.682 &0.671  \\ 
    SPV-MIA(Black Box/$D^{diff}$) &0.688  &0.676  &0.604  &0.583 &0.761 &0.689 \\  \bottomrule
    \end{tabular}}
    \caption{\label{lmia}The AUC of membership inference attacks on LLM under different threat models.}
    \vspace{-1.5em}
\end{table*}


\begin{table}[!h]
    \centering
    \small
    \scalebox{0.80}{\begin{tabular}{lccc}
    \toprule
    \multirow{2}{*}{Strategy} & \multicolumn{3}{c}{Text Generation Strategy} \\ \cmidrule{2-4}
    ~ & Top ($D^{no}$) & Temp ($D^{no}$) & Int ($D^{diff}$) \\
    \midrule
    Perplexity & 5 & 3 & 19 \\
    Small & 14 & 20 & 28 \\
    Medium & 18 & 10 & 22 \\
    zlib & 21 & 18 & 30 \\
    Window & 11 & 12 & 20 \\
    Lowercase & 24 & 10 & 30 \\ \hline
    Total Unique & 93 & 73 & 149 \\
    \bottomrule
    \end{tabular}}
    \caption{\label{tab:my_label}The number of memorized examples (out of 50
candidates) that we identify using each of the three text generation strategies and six membership inference techniques.}
\end{table}

\begin{table}[!t]
\small
  \centering
  \scalebox{0.80}{\begin{tabular}{c|cccc}
    \toprule
     Model& MRPC&QNLI&RTE&SST2\\ 
    \midrule
    Llama2 &0.743/0.576 & 0.910/0.768 &0.581/0.585 &0.905/0.890\\
    Qwen &0.743/0.777 & 0.910/0.768 &0.581/0.621 &0.905/0.894\\
  \bottomrule
\end{tabular}}
\caption{\label{llm-lmea-g}Results of the LMEA-G ($D^{sha}$). Original(BERT)/Acc.}
\end{table}

\begin{table}[!t]
  \centering
  \small
  \scalebox{0.90}{\begin{tabular}{c|cccc}
    \toprule
     \multirow{2}{*}{Model} & \multicolumn{2}{c}{CHIP-CTC} & \multicolumn{2}{c}{KUAKE-QIC}\\ \cmidrule{2-5}
     ~ &Original &ACC &Original &ACC\\
     \midrule
     Llama2 &0.771 & 0.765 &0.785 &0.778\\
     Qwen &0.782 & 0.770 &0.791 &0.790\\
  \bottomrule
\end{tabular}}
\caption{\label{llm-lmea-i}Results of the LMEA-I ($D^{sha}$, Lora).}
\vspace{-2.0em}
\end{table}

    \begin{table}[h]
    \centering
    \small
      \scalebox{0.90}{\begin{tabular}{c|cccc}
        \toprule
         \multirow{2}{*}{Model} & \multicolumn{2}{c}{CHIP-CTC} & \multicolumn{2}{c}{KUAKE-QIC}\\ \cmidrule{2-5}
         ~ &Original &ACC &Original &ACC\\
         \midrule
         Llama2 &0.771 & 0.768 &0.785 &0.774\\
         Qwen &0.782 & 0.780 &0.791 &0.790\\
      \bottomrule
    \end{tabular}}
    \caption{ \label{llm-lmea-i-3}Attack Results of the LMEA-I ($D^{par}$, Lora).}
    \end{table}

\begin{table}[!h]
\centering
\small
\scalebox{0.90}{\begin{tabular}{@{}lcccc@{}}
\toprule
& GPT2-Medium & GPT2-Large & GPT2-XL \\
\midrule
ECHR   & 3.38\%  & 17.37\%  & 14.21\%  \\
Enron   & 7.66\%  & 12.68\%  & 15.55\%  \\
\bottomrule
\end{tabular}}
\caption{\label{tab:ce} The attack success rate of LAIA (Black Box/Partial Data) on all datasets.}
\vspace{-1.5em}
\end{table}

\subsection{Experimental Results for NLP Models}
Due to space constraints, we only show results on Bert in the main text. The results for RoBert and GPT2 can be found in Appendix ~\ref{15}.

\textbf{Membership Inference Attack (MIA):} From Figure ~\ref{fig:mia_main}, we can draw the following conclusions: 1) The success rate of the MIA is positively correlated with the degree of overfitting. The degree of model overfitting can be seen in Tables ~\ref{tb1} and ~\ref{tb-ia} in the appendix ~\ref{13}. 2) White-box attacks' accuracy is better than black-box attacks', but the improvement is limited. 3) The accuracy of attacks in which the attacker has access to partial data is slightly higher than that of attacks using shadow data. 4) We find that the performance of the MIA and MEA are negatively correlated, a target model with higher MIA risks is less vulnerable to the MEA. For more details, refer to Appendix~\ref{15}.

\textbf{Model Inversion Attack (MDIA):} From Figure ~\ref{fig:mia_main}, we can draw the following conclusions: 1) The accuracy of MDIA is directly proportional to the complexity of the data. 
2) When the auxiliary data is labeled, the performance of the attack is significantly better than that when it is unlabeled. For more details, refer to Appendix~\ref{15}.
\vspace{-0.2em}
\textbf{Attribute Inference Attack (AIA):} From Figure ~\ref{fig:mia_main}, we can draw the following conclusions: 1) The success rate of attribute recognition for named entities will be higher than that for recognizing gender and age. 2) The performance of the white box is higher than that of the black box. For more details, refer to Appendix~\ref{15}.

\begin{table*}[ht]
\centering
\small
\scalebox{0.90}{\begin{tabular}{lccccccccc}
\toprule
Attr. & gender & location & married & age & education & occupation & pobp & income & Avg \\
\midrule
Acc. & 0.88 & 0.49 & 0.43 & 0.71 & 0.35 & 0.45 & 0.45 & 0.44 & 0.53 \\
\bottomrule
\end{tabular}}
\caption{\label{tab:attributes_accuracies} The attack success rate of LAIA (Black Box/No Data) on all attributes in the PR dataset (Llama2-7B).}
\end{table*}

\begin{table*}[ht]
\centering
\small
\scalebox{0.90}{\begin{tabular}{lccccccccc}
\toprule
Attr. & gender & location & married & age & education & occupation & pobp & income & Avg \\
\midrule
Acc. & 0.88 & 0.46 & 0.50 & 0.68 & 0.38 & 0.49 & 0.50 & 0.46 & 0.54 \\
\bottomrule
\end{tabular}}
\caption{\label{tab:attributes} The attack success rate of LAIA (Black Box/No Data) on all attributes in the PR dataset (Llama2-13B).}
\vspace{-1.5em}
\end{table*}

\textbf{Model Extraction Attack (MEA):} From Figure ~\ref{fig:mia_main}, we can draw the following conclusions: 1) MEAs generally can achieve high accuracy. 2) The performance using partial data is generally lower than using shadow data. 3) For models with overfitting, the experimental results tend to perform a little better when the temperature hyperparameter is lower. For more details, refer to Appendix~\ref{15}.

\subsection{Results of the Chained Framework} The data in Table ~\ref{tb7} leads us to deduce the subsequent conclusions: 1) The extraction model provides a strong defense against MIAs. This implies that if the model owner employs the MEA and publishes the extracted model, it could serve as an effective defense method against MIAs. 2) When an attacker (black-box) conducts an AIA on the extracted model in a white-box context, the attack performance is superior to that in a black-box scenario. 3) Leveraging the attack capability of the MDIA, we demonstrate a data-free MIA, whose accuracy is comparable to that of an attacker in possession of data with the same distribution. 4) We find that the MIA enhances the performance of the MEA on certain datasets (MRPC and RTE) when employed as a data filter. 5) Furthermore, on most datasets, data generated by the MDIA and filtered through the MIA can effectively increase the success rate of the MDIA. For more details, refer to Appendix~\ref{18}.

\subsection{Experimental Results for LLM Attacks}
\textbf{ Membership Inference Attack for LLM (LMIA):} From Table ~\ref{lmia}, we can draw the following conclusions: 1) Existed MIAs designed for LMs (based on overfitting) cannot handle LLMs with large-scale parameters. However, methods based on memorization are suitable for LLMs. 2) The LMIA does not necessarily require the use of shadow data; satisfactory attack performance can also be achieved with data from different domains. For more detailed results, refer to Appendix~\ref{19}.

\textbf{ Model Inversion Attack for LLM (LMDIA):} From Table ~\ref{tab:my_label}, we can draw the following conclusions: 1) The LMDIA is more inclined to recover news headlines and log files. 2) The attacker's success rate using cross-domain data is higher than that of the no-data assumption. For more detailed results, refer to Appendix~\ref{19}.

\textbf{ Attribute Inference Attack for LLM (LAIA):} From Table ~\ref{tab:attributes},~\ref{tab:attributes_accuracies} and ~\ref{tab:ce}, we can draw the following conclusions: 1) The capability of LAIA is correlated with the size of the model; the larger the model, the higher the success rate of the attribute inference attacks. 2) Age and gender are attributes that are easier to infer, whereas education and occupation are more difficult to infer. For more details, refer to Appendix~\ref{19}.

\textbf{ Model Extraction Attack for LLM (LMEA)}: From Table ~\ref{llm-lmea-g},~\ref{llm-lmea-i} and ~\ref{llm-lmea-i-3}, we can draw the following conclusions: 1) Compared to llama2, Qwen is more vulnerable to the LMEA attack, consequently revealing more of its knowledge. 2) Regardless of the threat model and fine-tuning strategy employed, attackers can easily transfer knowledge from the target model to the extraction model. 3) General LLMs possess a stronger capacity to resist LMEA compared to domain-specific LLMs. For more detailed results, refer to Appendix~\ref{19}.

\section{Related Work}
\textbf{Privacy Attacks}: Shokri et al.~\cite{Shokri2016MembershipIA} present the first MIA on ML models in the black-box case: it trains multiple shadow models to simulate the behavior of the target model and uses multiple classification models to distinguish members and non-members.
Fredrikson et al.~\cite{fredrikson2014privacy} is the first to propose model inversion attacks on classification models. Subsequently, they use back-propagation of gradients to recover face information in ~\cite{Fredrikson2015ModelIA}. 
Song et al.~\cite{iclr/SongS20} propose the use of classification models to improve the performance of attribute inference attacks and expose the link between overfitting and attack performance.
Tramèr et al.~\cite{tramer2016stealing} propose the first model extraction attack against machine learning APIs. 
Jia et al.~\cite{jia2019memguard} observe that when the attack model is a binary classifier, it is vulnerable to adversarial examples. 

\section{Conclusion}
In this paper, we develop a privacy attack and defense system for NLP models (the conventional/small models and LLMs). 
To be more realistic, we have done extensive experiments with auxiliary data from different domains.
We further use KD to mitigate the poor performance of MIAs. On the other hand, we propose a chained framework to enable a practitioner to chain multiple attacks to achieve a
higher-level objective. In real-world applications, our system can do a comprehensive privacy evaluation for NLP models to enable users to fully understand the extent of the model's leak privacy before it is deployed. 

\section{Limitations}
In this paper, we introduce a benchmark for privacy evaluation in NLP, incorporating a wide array of methods tailored to different threat models. Throughout our research and implementation, we have uncovered several unexplored areas in NLP privacy attacks, including a shortage of black-box model inversion attacks on small NLP models and the insufficient accuracy of white-box model inversion attacks. We are committed to advancing the research in this field. Additionally, we propose a chain framework in this paper. While we have identified 14 types of connections, we believe there are many more potential connections between different attacks that could achieve higher attack objectives. To date, our research has focused on pairwise attack connections, but we intend to investigate more complex connections, including those involving three or more than three attacks. Although Large Language Models (LLMs) are in their nascent stages and evolving rapidly, research on privacy attacks for these models is not as developed as for smaller models. We plan to prioritize privacy attacks on LLMs in our future work, aiming to contribute to the development of the privacy community surrounding LLMs.


\bibliography{acl_latex}

\appendix
\section{How to Use Benchmark}
\label{htub}

\begin{figure*}[!ht]
    \begin{center}       \includegraphics[width=0.90\linewidth]{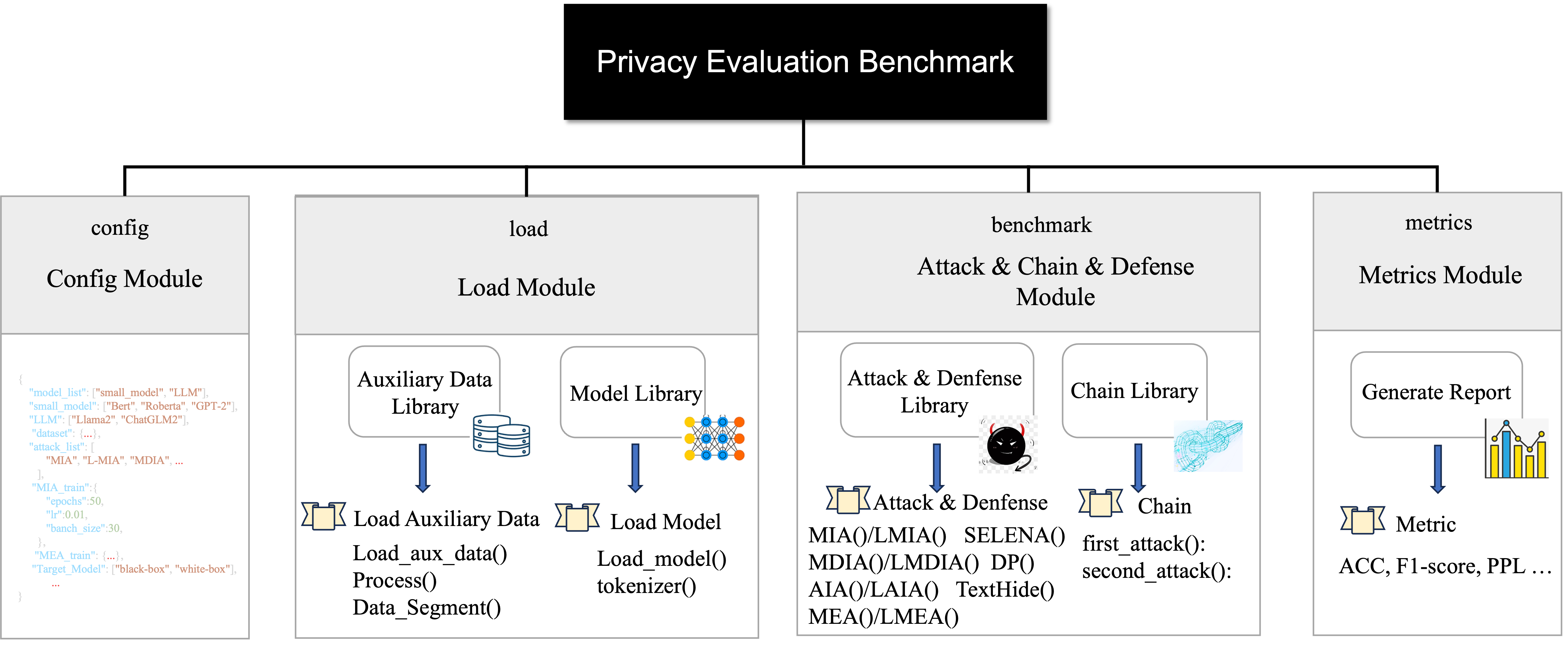}
    \end{center}
    \vspace{-0.3cm}
    \caption{Workflow of benchmark}
    \label{fig:train}
    \vspace{-1.2em}
    \end{figure*}
The following steps are required to use the privacy evaluation benchmark:

\textbf{Build:} Download our code repository from GitHub and prepare all the required environments.

\textbf{Configure:} Modify the configuration JSON file to specify settings, including the model, auxiliary data, attack type, defense method, target model, and attack model training hyperparameters (such as 'epoch' and 'lr', etc.). Descriptions of all hyperparameters can be found in their respective folders, as seen in the README.md located in each folder.   

\textbf{Load Model and Data:} Users can use Load\_model() and Load\_aux\_data() to handle the uploaded model and auxiliary data, and use tokenizer() to encode texts. We leverage Hugging Face's \cite{wolf2020transformers} open-source libraries to load models and data.

\textbf{Attack \& Chain \& Defense:}  Users can navigate to the relevant folder as specified by the attack and defense types in the configuration file and execute shell scripts to run the attack and defense mechanisms. To operate within the chained framework, preliminary attacks must be conducted under the designated attack and threat models before proceeding with the follow-up attacks.

\textbf{Evaluate:} User can use compute\_metrics() in run\_glue.py to obtain results and evaluations.   
We summarize the above steps in Figure ~\ref{fig:train}. Before deploying the target model in the real world, model owners can assess the privacy threats their models may face from our benchmark.

\section{Privacy Attacks for NLP Models}
\label{1}
This paper will focus on four privacy attacks against NLP small models: MIA, MDIA, AIA, and MEA. MIA is to infer whether the target sample is included
in the training data. MDIA is to reconstruct the original training samples. AIA obtains information about attributes of the training data that are not relevant to the original task. MEA is used to reconstruct the model parameters or model functions.

\subsection{Membership Inference Attack (MIA)} 
\label{2}
The goal of MIA is to infer whether the target sample is included in the training data of the target model. Inferring the membership of the target sample can trigger a serious user privacy breach. For example, if the target model is trained using the user's bank transaction records, MIA will leak the user's financial and transaction details. 
Next, we describe how an attacker can execute a membership inference attack under different assumptions.

\textbf{Black Box (Threshold or Classification Model based Methods)/(Shadow Data or Partial Data)~\cite{shejwalkar2021membership,47e5,song2021systematic}}: We introduce two approaches for the black-box MIAs. The former distinguishes members and non-members based on a threshold~\cite{shejwalkar2021membership,song2021systematic}, and the latter trains a classification model for this purpose~\cite{47e5}. In the first step, the attacker uses the shadow training data and shadow testing data to query the shadow model (the shadow model has been trained using the shadow training data) to get the attack feature (the first method uses losses obtained by comparing predictions with labels, the second method uses logits). In the second step, an adversary uses the attack feature obtained in the previous step to calculate the threshold for different labels (the loss of the member data is less than this threshold and vice versa) or to train the attack model (the label of the member data is 1 and the opposite is 0). In the third step, the target samples are fed to target model to obtain losses or logits, and finally the threshold or the attack model obtained in previous steps are used for judgment. In this scenario, the auxiliary data the attacker used are the shadow data or partial data. The difference is that the attacker in the case using the partial data does not need to train a shadow model, but directly queries the target model.

\textbf{White Box (Classification Model based Methods)/(Shadow Data or Partial Data)~\cite{sp/NasrSH19}}: The white-box MIA differs from the black-box one in that the attacker can access to all information of the model, so the attacker can train the attack model using gradients, outputs of the last activation layer, logits, classification losses and labels of the target model. The rest of steps are similar to that of the black-box attack.

\subsection{Model Inversion Attack (MDIA)} 
\label{3}
The goal of MDIA is to reconstruct the original training samples, and there is some work in the field of image classification. It is difficult to implement this attack in NLP since the text is the serialized data. So we referring to the image MDIA methods.

\textbf{White Box/(No Data)~\cite{Fredrikson2015ModelIA}}: This approach assumes that the attacker can obtain all knowledge of the model and does not require any auxiliary data. First, a batch of noisy samples is generated and labels are randomly assigned  to them. Then, the samples are fed into the target model to obtain losses, followed by updating this sample (here the sample serves as learnable parameters) using the gradient descent algorithm until the classification loss is less than a threshold set in advance or the number of iteration rounds is reached.

\textbf{White Box/(Shadow Data)~\cite{zhang2022text,iclr/DathathriMLHFMY20}}: A model inversion attack in the field of NLP classification was first proposed by Ruisi Zhang et al. First, This work starts with an N-gram analysis using the shadow data to get the prompt texts for subsequent generation, called templates (that paper considered the auxiliary data to be unlabeled data, The thesis also takes into account the scenario with labels.). Then it uses the shadow data to fine-tune GPT2. Finally it applies feedback from the target model to perturb the hidden state of the GPT2 model to generate texts.
\begin{table*}[!ht]
    \centering
    \small
   \scalebox{0.77}{
    \begin{tabular}{ccccccccccc}
    \toprule
    \multicolumn{2}{c}{} & \multicolumn{2}{c}{Target Model} & \multicolumn{4}{c}{Auxiliary Data} & \multicolumn{3}{c}{Attacker's Objective} \\ \cmidrule{3-11}
         \multicolumn{2}{c}{}      & $B^{box}$         & $W^{box}$         & $D^{sha}$ & $D^{par}$ & $D^{no}$ & $D^{diff}$ & Training Data & Attributes &Parameters \\ \midrule
         \multicolumn{11}{c}{Conventional/Small Model (BERT~\cite{devlin2018bert}, RoBERTa~\cite{liu2019roberta}, GPT2-small~\cite{radford2019language})}           \\
         ~ &$<$MIA, $B^{box}$, $D^{sha}$$>$ &\checkmark &- &\checkmark &- &- &- &\checkmark &- &-  \\
         ~ &$<$MIA, $B^{box}$, $D^{par}$$>$ &\checkmark &- &- &\checkmark &- &- &\checkmark &- &-  \\
         ~ &$<$MIA, $B^{box}$, $D^{diff}$$>$ &\checkmark &- &- &- &- &\checkmark &\checkmark &- &-  \\
         ~ &$<$MIA, $W^{box}$, $D^{sha}$$>$ &- &\checkmark &\checkmark &- &- &- &\checkmark &- &-  \\
         ~ &$<$MIA, $W^{box}$, $D^{par}$$>$ &- &\checkmark &- &\checkmark &- &- &\checkmark &- &-  \\
         ~ &$<$MIA, $W^{box}$, $D^{diff}$$>$ &- &\checkmark &- &- &- &\checkmark &\checkmark &- &-  \\
         ~ &$<$MDIA, $W^{box}$, $D^{sha}$$>$ &- &\checkmark &\checkmark &- &- &- &\checkmark &- &-  \\
         ~ &$<$MDIA, $W^{box}$, $D^{no}$$>$ &- &\checkmark &- &- &\checkmark &- &\checkmark &- &-  \\
         ~ &$<$MDIA, $W^{box}$, $D^{diff}$$>$ &- &\checkmark &- &- &- &\checkmark &\checkmark &- &-  \\
         ~ &$<$AIA, $B^{box}$ , $D^{sha}$$>$ &\checkmark &- &\checkmark &- &- &- &- &\checkmark &- \\
         ~ &$<$AIA, $W^{box}$, $D^{sha}$$>$ &- &\checkmark &\checkmark &- &- &- &- &\checkmark &-  \\
         ~ &$<$MEA, $B^{box}$, $D^{sha}$$>$ &\checkmark & &\checkmark &- &- &- &- &- &\checkmark  \\
         ~ &$<$MEA, $B^{box}$, $D^{par}$$>$ &- &\checkmark &- &- &\checkmark &- &- &- &\checkmark  \\
         ~ &$<$MEA, $B^{box}$, $D^{diff}$$>$ &- &\checkmark &- &- &- &\checkmark &- &- &\checkmark  \\
         \multicolumn{11}{c}{LLM (Llama2~\cite{touvron2023llama},Qwen~\cite{bai2023qwen},GPT2-xl~\cite{radford2019language})}           \\
         ~ &$<$LMIA, $B^{box}$, $D^{sha}$$>$ &\checkmark &- &\checkmark &- &- &- &\checkmark &- &-  \\
         ~ &$<$LMIA, $B^{box}$, $D^{diff}$$>$ &\checkmark &- &- &- &- &\checkmark &\checkmark & -&-  \\
         ~ &$<$LMDIA, $B^{box}$, $D^{no}$$>$ &\checkmark &- &- &- &\checkmark &- &\checkmark &- &-  \\
         ~ &$<$LMDIA, $B^{box}$, $D^{diff}$$>$ &\checkmark &- &- &- &- &\checkmark &\checkmark &- &-  \\
         ~ &$<$LAIA, $B^{box}$, $D^{no}$$>$ &\checkmark &- &- &- &\checkmark &- &- &\checkmark &-  \\
         ~ &$<$LAIA, $B^{box}$, $D^{par}$$>$ &\checkmark &- &- &\checkmark &- &- &- &\checkmark & - \\
         ~ &$<$LMEA, $B^{box}$, $D^{sha}$$>$ &\checkmark &- &\checkmark &- &- &- &- &- &\checkmark  \\
         ~ &$<$LMEA, $B^{box}$, $D^{par}$$>$ &\checkmark &- &- &\checkmark &- &- &- &- &\checkmark  \\
          \bottomrule
\end{tabular}}
\caption{\label{ttt}An overview of the characteristics of different attacks and their corresponding threat models. \checkmark indicates the attack has this feature, – indicates the attack does not have this feature.}
\vspace{-1.2em}
\end{table*}

\subsection{Attribute Inference Attack}
\label{4}
The goal of AIA is to infer attribute information of the target data. During the training process, the target model unintentionally learns additional information in addition to the classification information of the original task. In this case, the attacker can obtain extra attribute information (age, gender, name entity, etc.) of the corresponding sample. 

\textbf{Black Box and White Box/(shadow data)~\cite{Lyu2021KillingTB,iclr/SongS20}}: An attacker uses the shadow data to query the target model to obtain model outputs (logits for the black-box case and outputs of encoders for the white-boxe case) and uses attributes as labels to obtain the attack features. Next, the attack features are used to train the attack model. Finally, the target samples are fed into the target model, and the target model outputs predictions.

\subsection{Model Extraction Attack}
\label{5}
The goal of MEA is to reconstruct the parameters of the target model, so that the attacker can obtain an extracted model with a comparable performance to the target model. MEA is mainly based on the scenarios where the target model is accessed in the form of APIs. 

\textbf{Black Box/ (Shadow Data or Partial Data)~\cite{Lyu2021KillingTB,he2021model}}: The attacker queries the target model with the shadow data (unlabelled data) to obtain logits as soft labels, and then uses the shadow data and soft labels to train a function-extracted model.

\subsection{Privacy Attacks using Auxiliary Data from Different Domains}
\label{6}
To be more practical, we remove the assumption that the attacker has data from the same distribution, and instead that the adversary has data from different domains. 

The attack strategy under data from different domains is very similar to previous privacy attacks. The difference is that the attacker uses an NLP dataset with different distributions from Target Training Data as an auxiliary dataset. We have conducted numerous experiments on MIAs, MIDAs, and MEAs (AIA are not taken into account here, because the attribute information varies across domains, which makes it difficult to obtain meaningful results). 

\textbf{Attack Strategy of Membership Inference}: The attacker uses data from different domains to train the shadow model. 

\textbf{Attack Strategy of Model Inverse}: 
The hacker under this attack uses data from different domains to query the target model to obtain the output of the embedding layer, and later uses this output as the initial sample for the attacker. 

\textbf{Attack Strategy of Model Extraction}: In this scenario, we assume that the data obtained from different domains are unlabeled, and we query the target model for labeling to obtain soft labels.

\subsection{Privacy Defenses for NLP Models}
\label{7}
From the previous introduction, we have learned that privacy attacks can cause serious damage, so defenses against them are essential. Here, we select three classical or SOTA methods, namely, DP-SGD~\cite{abadi2016deep}, SELENA~\cite{280000}, and TextHide~\cite{huang2020texthide}. specifically, DP-SGD adds Gaussian noise to the gradient during the model training. SELENA is a framework to train privacy-preserving models that induce similar behaviors on member and non-member inputs to mitigate MIA. Texthide is a text privacy protection technique and requires each participant to add a simple step to hide the representation of their text data with a one-time encryption.

\section{Attacks for LLM}
\label{8}
In this study, we focus on four LLM privacy attacks, namely LMIA, LMDIA, LAIA, and LMEA.

\subsection{Membership Inference Attack for LLM 
 (LMIA):} 
\label{9}
\textbf{Black Box/(Shaow Data)~\cite{mireshghallah2022empirical}:} We introduced a reference based attack, which adopts the pre-trained model as the reference model to calibrate the likelihood metric to infer membership (denoted as LiRA).

\textbf{Black Box/(Data from Different Domains)~\cite{mattern2023membership,fu2023practical}:} In practical scenarios, assuming that the attacker has access to shadow data is a strong assumption. To better fit real-world situations, Mattern et al~\cite{mattern2023membership}, propose neighborhood attacks (Neighbour Attack), which compare model scores for a given sample to scores of synthetically generated neighbor texts and therefore eliminate the need for access to the training data distribution. For details, refer to Equation ~\ref{equ}, where $f_{\theta}$ represents the target model, $x$ denotes the target sample point, $\tilde{x}_i$ signifies the neighboring sample point (obtained by replacing words in $x$), and $\gamma$ indicates the threshold value. 
\begin{equation}
\label{equ}
A_{f_{\theta}}(x) = \mathbbm{1}[{L(f_{\theta}, x)} - \frac{\sum_{i=1}^{n} L(f_{\theta}, \tilde{x}_i)}{n} < \gamma ]
\end{equation}

Fu et al~\cite{fu2023practical}, propose a membership inference attack based on Self-calibrated Probabilistic Variation (SPV-MIA). Specifically, recognizing that memorization in LLMs is inevitable during the training process and occurs before overfitting, they introduce a more reliable membership signal, probabilistic variation, which is based on memorization rather than overfitting. Furthermore, they introduce a self-prompt approach, which constructs the dataset to fine-tune the reference model by prompting the target LLM itself. For details, refer to Equation ~\ref{equ1}, where $\tilde p_{\theta}(x)$
and $\tilde p_{\phi} (x)$ are probabilistic variations of the text record $x$ measured on the target model $\theta$ and the reference model $\phi$ respectively.
\begin{equation}
\label{equ1}
A_{our} (x, \theta, \phi) = \mathbbm{1}[\tilde p_{\theta}(x) - \tilde p_{\phi} (x) \leq \gamma]
\end{equation}
\subsection{ Model Inversion Attack for LLM (LMDIA):} 
\label{10}
\textbf{Black Box/(No Data/ Data from Different Domains)~\cite{274574}:} We introduce one approach for the black-box.
This work first builds three datasets of 200,000 generated samples (each of which is 256 tokens long) using one of the following strategies:
\begin{itemize}
 \item Top-n (Top): samples naively from the empty sequence.
 \item Temperature (Temp): increases diversity during sampling.
 \item  Internet (Int): conditions the LM on Internet text.
\end{itemize}
We order each of these three datasets according to each of the six membership inference metrics:
\begin{itemize}
 \item Perplexity: the perplexity of the largest GPT-2 model.
 \item Small: the ratio of log-perplexities of the largest GPT-2
model and the Small GPT-2 model.
 \item Medium: the ratio as above, but for the Medium GPT-2.
 \item zlib: the ratio of the (log) of the GPT-2 perplexity and the zlib entropy (as computed by compressing the text).
 \item Lowercase: the ratio of perplexities of the GPT-2 model on the original sample and on the lowercased sample.
 \item Window: the minimum perplexity of the largest GPT-2 model across any sliding window of 50 tokens.
\end{itemize}
For each of these 3×6 = 18 configurations, we select 100 samples from among the top 1000 samples. 
\begin{table*}[!t]
    \centering
    \small
    \scalebox{0.90}{\begin{tabular}{c|cccccc}
    \toprule
        Dataset & \#Label & Attribute & \#Train & \#Dev & Task & Source \\  \midrule
        MRPC & 2 & $\backslash$ & 3668 & 408 & paraphrase & news \\ 
        QNLI & 2 & $\backslash$ & 103k & 5K & QA/NLI & Wikipedia \\ 
        RTE & 2 & $\backslash$ & 2490 & 277 & NLI & news/Wikipedia \\ 
        AG\_News & 4 & $\backslash$ & 120K & 7.6K & Topic Classification & news \\ 
        SST2 & 2 & $\backslash$ & 67K & 872 & sentiment & movie reviews \\ 
        TREC & 6 & $\backslash$ & 5000 & 452 & Topic Classification & QA \\ 
        YELP\_Polairty & 2 & $\backslash$ & 560K & 38K & sentiment & movie reviews \\ 
        TP & 5 & age, gender & 24K & 2767 & sentiment & Trustpilot Sentiment dataset \\ 
        AG\_News(AIA) & 4 & entity & 13K & 1457 & Topic Classification & news \\ 
        BLOG & 10 & age, gender & 7985 & 887 & Topic Classification & blog authorship corpus \\ 
        CHIP-CTC &44 & $\backslash$ & 22k & 7k & Medical & Clinical trial \\
        KUAKE-QTR &11 & $\backslash$ & 6931 & 2K & Medical & Medical search \\
        KUAKE-IR &2 & $\backslash$ & 5000 & 600 & Medical & Medical search \\
        ECHR & 1  & name & 7.1k & 1.38k & Legal & law cases dealt \\
        Enron & 4 &  e-mail, work address.. & 31.7k & 2K & e-mail & e-mail \\ 
        PersonalReddit & 8 &  age, education.. & 1184 & - & Reddit & Reddit profiles \\
        \bottomrule
    \end{tabular}}
    \caption{Dataset introduction}
    \label{data}
\end{table*}
\subsection{ Attribute Inference Attack for LLM (LAIA):} 
\label{11}
\textbf{Black Box/(No Data or Partial Data)~\cite{staab2023beyond, lukas2023analyzing}:} Recently, Staab et al. proposed a method for an LLM attribute inference attack. Specifically, Given $t$ (text), $A_1$ (Adversary) first creates a prompt $P_{A_1}(t) = (S, P)$. For this, $P_{A_1}$ is a function that takes in the text $t$ and produces both a system prompt $S$ and a prompt $P$ which is given to the model $M$. While this formulation is general, for the rest of this work, they restrict the prompt $P$ to $P = (Prefix{F_{A_1}}(t)Suffix)$ where $F_{A_1}$ is a string formatting function. The specific form of the prompt can be seen in the original paper. The model $M$ responds to this prompt with $M(P_{A_1}(t)) = {(a_j , v_j )}_{1 \leq j \leq k}$ (Where 'a' represents attribute, 'v' represents value.) the set of tuples it could infer from the text. 

Lukas et al. proposed a PII (Personally Identifiable Information) reconstruction, they assume a more informed attacker, similar to that of membership inference, who has some knowledge about the dataset. For example, when an attacker wants to learn more PII about a user, they can form masked queries (e.g., "John Doe lives in [MASK], England") to the LM and attempt to reconstruct the missing PII.

\subsection{ Model Extraction Attack for LLM (LMEA):}
\label{12}
\textbf{Black Box/(Shaow Data or Partial Data):} This paper employs two attacks suitable for black-box scenarios. The first approach denoted as LMEA-G ($D^{sha}$ or $D^{par}$), targets generalized LLMs ~\cite{tang2023does}. It involves using an LLM to annotate unlabeled data, which is then used to train smaller models like BERT.

The second approach, LMEA-I ($D^{sha}$ or $D^{par}$) ~\cite{gu2023knowledge}, is tailored for industrial LLMs. We utilize strategies for parameter-efficient Tuning, such as LoRA ~\cite{hu2021lora} and P-Tuning v2 ~\cite{liu2021p}. This method relies on shadow or partial data to extract soft labels that are subsequently used to train the extraction model.

\section{Results of Attacks and Defenses}
\label{13}
In this section, we describe experimental data, the model, and the results of experiments on NLP privacy attacks and defenses.

\subsection{Datasets, Models, and Settings}
\label{14}
In this paper, we selected fifteen experimental datasets.  
They are: MRPC~\cite{dolan2005automatically}, RTE\cite{wang2019glue}, YELP\_Polarity~\cite{Zhang2015CharacterlevelCN}, AG\_News~\cite{Zhang2015CharacterlevelCN}, TP~\cite{coavoux2018privacy}, BLOG~\cite{Schler2006EffectsOA}, SST2~\cite{socher-etal-2013-recursive}, QNLI~\cite{wang2019glue}, TREC~\cite{li2002learning}, CHIP-CTC~\cite{zhang2021cblue}, KUAKE-QIC~\cite{zhang2021cblue}, Wikitext-103 (Wiki)~\cite{merity2016pointer}, ECHR~\cite{chalkidis2019neural}, Enron~\cite{klimt2004introducing} and PersonalReddit (PR)~\cite{staab2023beyond} (CHIP-CTC and KUAKE-QIC are used for LMEA experiments. ECHR, Enrom and PR are used for LAIA experiments). 
Because attribute inference attacks require data to have attribute labels, we introduced five additional datasets in this attack (TP, BLOG, ECHR, Enron and PR), and used the remaining datasets as supplementary data in experiments across different domains. The specifics of the data can be found in ~\ref{data}.

\begin{table}[!t]
  \centering
  \small
  \scalebox{0.90}{\begin{tabular}{c|ccc}
  \hline
  ~ & BERT & RoBERTa & GPT2 \\
  \hline
  MRPC & 0.9640/0.7426 & 0.8974/0.7966 & 0.8337/0.7353 \\
  RTE & 0.9775/0.5812 & 0.9389/0.6173 & 0.9695/0.5993 \\
  AG\_News & 0.9897/0.9243 & 0.9838/0.9353 & 0.9872/0.9318 \\
  YELP & 0.9984/0.9498 & 0.9975/0.9603 & 0.9977/0.9585 \\
  \hline
  \end{tabular}}
  \caption{Performance of the target model on training/testing sets on each dataset.}
  \label{tb1}
\end{table}

\begin{table}[!t]
    \centering
    \small
    \scalebox{0.90}{\begin{tabular}{c|ccc}
    \toprule
        ~ & AG(AIA) & BLOG & TP \\ \midrule
        BERT & 0.7934 & 0.9391 & 0.8652 \\ 
        RoBERTa & 0.8058 & 0.9481 & 0.8699 \\ 
        GPT2 & 0.7907 & 0.9200 & 0.8771 \\ \bottomrule
    \end{tabular}}
    \caption{Inference accuracy of different models using different datasets.}
    \label{tb-ia}
\end{table}

To verify the effectiveness of different attack methods under different models, six commonly used NLP models, namely BERT~\cite{devlin2018bert}, RoBERTa~\cite{liu2019roberta}, GPT2-small (subsequently abbreviated as GPT2)~\cite{radford2019language}, Qwen-7B~\cite{zeng2023glm-130b}, Llama2 ~\cite{touvron2023llama} and GPT2-xl~\cite{radford2019language} were chosen for this study. The experimental setup can be seen in the description below, and the performance of the target model on each dataset is shown in Table ~\ref{tb1} and ~\ref{tb-ia}. 

\textbf{Performance Metrics}
We use four metrics to measure Attack Performance and Defense Performance, namely Accuracy (the proportion of correctly predicted samples to the total number of samples, with higher values indicating better attack performance), F1-score (a composite of precision and recall, with higher values indicating better attack performance), PPL (Perplexity, used to reflect the fluency of the text, with lower values indicating better model counter-attack effectiveness), and AUC (Area Under the ROC Curve, is a metric used to measure the performance of binary classification models. The closer the AUC value is to 1, the better the model is at distinguishing between positive and negative classes. An AUC value of 0.5 indicates that the model has no discriminative power, equivalent to random guessing.).

\textbf{Target model training.} In this paper, we chose to use bert-base-uncased as the BERT model, for the MRPC dataset we used a learning rate of 2e-5 and batch\_size of 32. We completed all experiments on the open-source framework transformers.

\textbf{Membership Inference Attack (Settings).} We train the shadow model in the same way in different membership inference attacks, where we ensure that the architecture and training process of the shadow model and the hyperparameters chosen are consistent with the target model. In the approach of the classification model as an attack model, we construct different linear layers for different information. In this case, one linear layer was used for the loss values, and 3 layers were used for the logits, for the gradient information we chose to go through one CNN layer and then pick up 2 linear layers, and 2 linear layers were used for the labels. Finally, we combined the outputs of the different linear layers and fed them to the four linear layers. We use RELU as the activation function for the attack model, batch\_size is set to 32, Adam as the optimizer, cross-entropy as the loss function and the learning rate is set to 1e-5.

\textbf{Model Inversion Attack (Settings).} In this paper, there are two different model inversion attacks. For the first model inversion attack, we set alpha as 50, beta as 20, gama as 0.001, learning\_rate as 0.1, and set the number of recovered embeddings as 100. For the second model inversion attack, n was set to [4,7) and [6,8) for AG\_News and YELP\_Polarity, respectively, when doing the n-gram analysis. For training, we set kl\_loss to 0, window\_length to 0, step to 0.004, and set the number of repetitions to 10. In the evaluation phase, we selected the evaluation models Distil-Bert, Distil-RoBerta, and Distil-GPT2 and followed the transformers' script for training.

\textbf{Attribute Inference Attack (Settings).} For the attribute inference attack, we used 4 linear layers in training the attack model and set the learning rate as 1e-4, the optimizer as Adam, the batch\_size as 32, and the loss function as the cross-entropy loss function.

\textbf{Model Extraction Attack (Settings).} When training the extraction model in this paper, the logits obtained by querying the target model with shadow data are used as soft labels. We ensure that the architecture and training process of the extraction model is consistent with the target model.
\begin{figure*}[!h]
\begin{center}
    \includegraphics[width=0.8\linewidth]{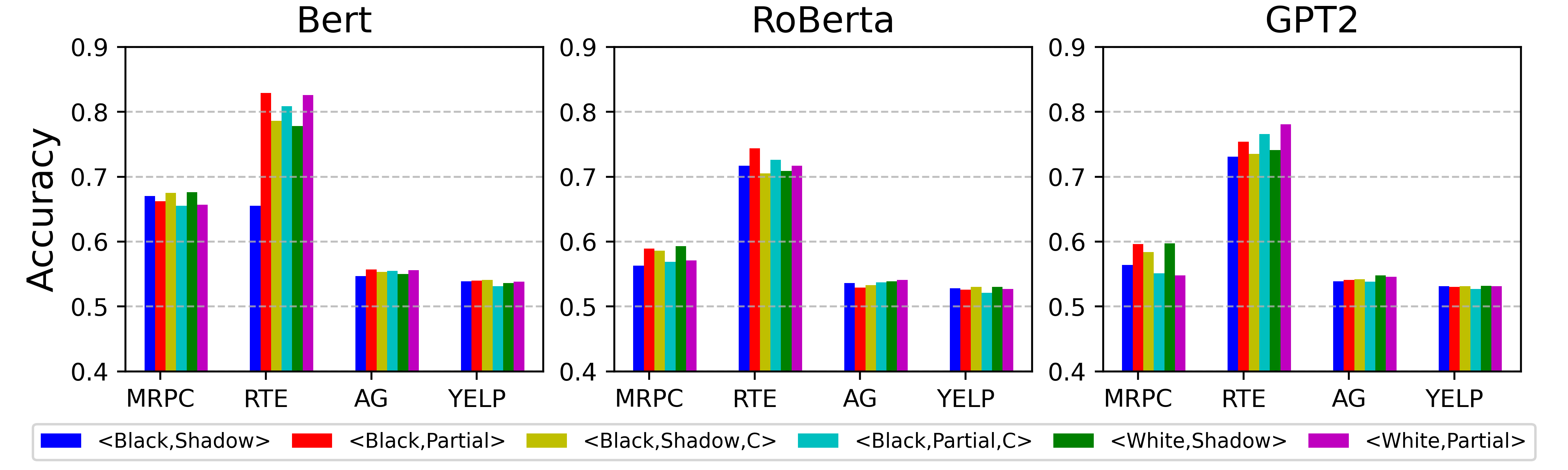}
\end{center}
\setlength{\abovecaptionskip}{-0.2em}
\caption{\label{fig:mia}Attack accuracy of MIA under different data and model architectures.}
\vspace{-0.1cm}
\end{figure*}
\begin{figure*}[!h]
\begin{center}
    \includegraphics[width=0.8\linewidth]{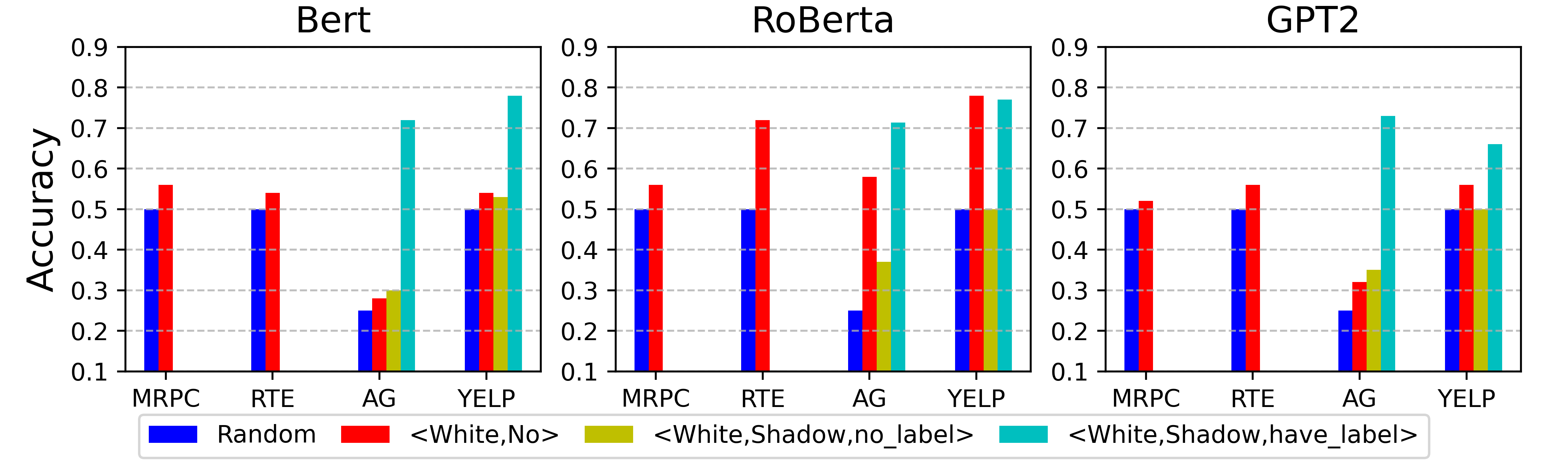}
\end{center}
\setlength{\abovecaptionskip}{-0.2em}
\caption{\label{fig:mdia}Attack accuracy of MDIA under different data and model architectures.}
\vspace{-0.1cm}
\end{figure*}
\begin{figure*}[!h]
\begin{center}
    \includegraphics[width=0.8\linewidth]{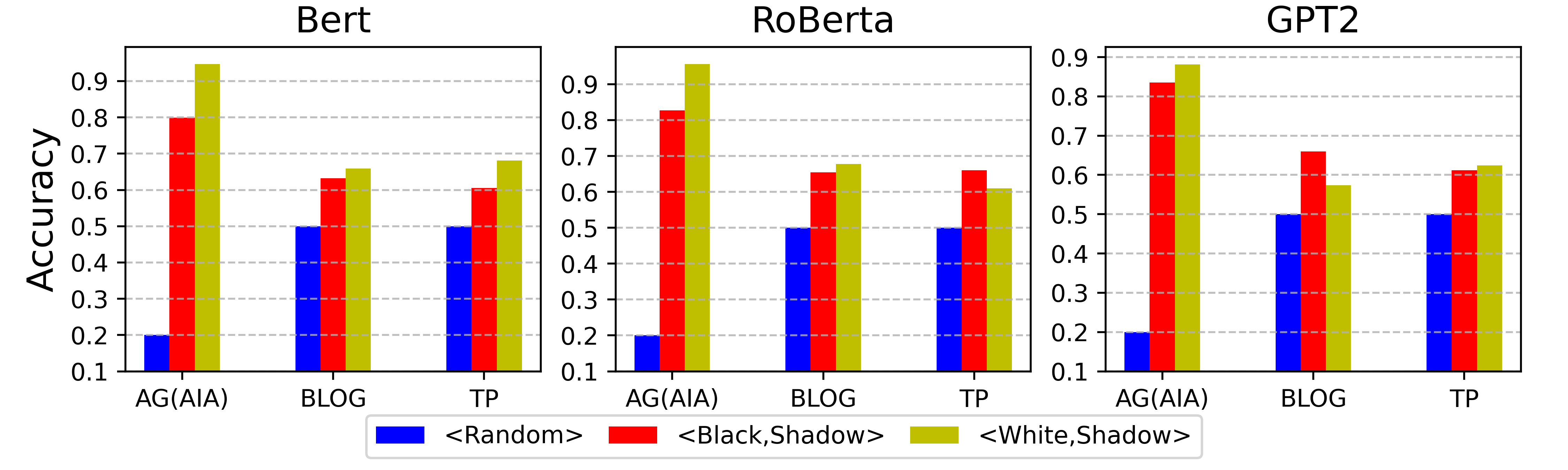}
\end{center}
\setlength{\abovecaptionskip}{-0.2em}
\caption{\label{fig:aia}Attack accuracy of AIA under different data and model architectures.}
\vspace{-0.1cm}
\end{figure*}
\begin{figure*}[!h]
\begin{center}
    \includegraphics[width=0.8\linewidth]{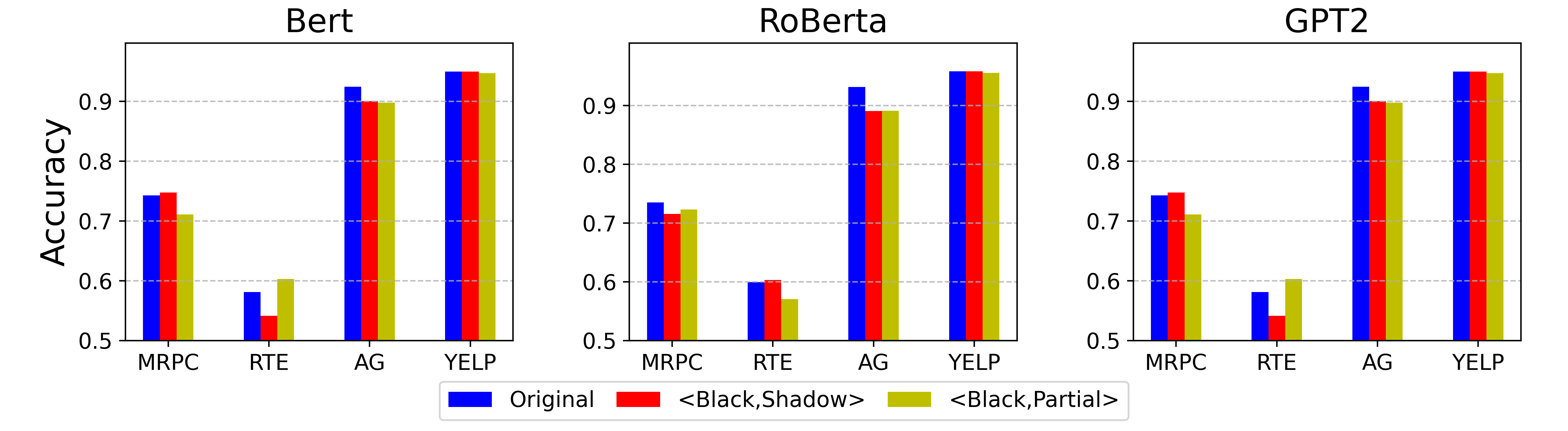}
\end{center}
\setlength{\abovecaptionskip}{-0.2em}
\caption{\label{fig:mea}Attack accuracy of MEA under different data and model architectures.}
\end{figure*}

\subsection{Privacy Attack Performance}
\label{15}

\textbf{Membership Inference Attack (MIA)}: In Figure~\ref{fig:mia}, we report the accuracy of MIA. We observe that the attack achieves high accuracy on both MRPC and RTE, for example, the BERT model trained on RTE dataset has a success rate of 0.778 for the attacker under the white-box assumption(shadow data). On the one hand, the attack performance on AG\_News and YELP\_Polarity is relatively low, mainly because these two datasets have better generalization performance, i.e. the model is less overfitted.

As we can see from Figure~\ref{fig:mia}, when the attacker obtains the target model as white box, the success rate of attack is better than black box in most cases, although the difference is not substantial. Secondly, when the attacker obtains partial data, the attack performance is slightly better than shadow data, but also remains at the same level. We also verified that using the threshold approach or classification model approach had little effect on attack performance.

\textbf{Model Inversion Attack (MDIA)}: The accuracy of MDIA can be seen in Figure~\ref{fig:mdia}. For the second attack, we only report performance on AG\_News and YELP\_Polarity, as the method can only recover single-sentence data. From Figure~\ref{fig:mdia} we can see that BERT and GPT2 are not effective compared to RoBERTa in the MDIA ($D^{no}$). the success rate of the former two on RTE is around 0.55 compared to 0.72 for the latter, mainly because RoBERTa model learns more information about the data. On the second point, the performance of the attack is significantly better when the auxiliary data is labeled than when it is unlabeled. For example, when the target model is BERT trained on the YELP\_Polarity dataset, the attack performance is 0.78 in the labeled scenario, compared to 0.53 in the unlabeled scenario.

\textbf{Attribute Inference Attack (AIA)}: We measured the performance of AIA, and present the results in Figure~\ref{fig:aia}. From figure, we can see that the success rate of attack is higher on AG\_News (AIA) and relatively lower on BLOG and TP. One reason for this may be that the former's attribute is name entity, while the latter two were gender and age, and name entity is more likely to be learned by model. Secondly, we also observed that the white box's performance is higher than the black box, suggesting that the representation of the model contains more information about attributes. We also report metrics for the F1-score, which can be found in Table ~\ref{aiaf1}.

\begin{table}[!t]
    \centering
    \small
    \scalebox{0.90}{\begin{tabular}{c|ccc}
    \toprule
        ~ & AG(AIA) & BLOG & TP \\ \midrule
        $<$BERT,Black,Shadow$>$ & 0.4212 & 0.544 & 0.6086 \\ 
        $<$BERT,White,Shadow$>$ & 0.8850 & 0.6915 & 0.7062 \\ \hline
        $<$RoBERTa,Black,Shadow$>$ & 0.5258 & 0.7049 & 0.6485 \\ 
        $<$RoBERTa,White,Shadow$>$ & 0.9099 & 0.7171 & 0.6125 \\ \hline
        $<$GPT2,Black,Shadow$>$ & 0.5659 & 0.6955 & 0.6015 \\ 
        $<$GPT2,White,Shadow$>$ & 0.7259 & 0.4952 & 0.6170 \\ \bottomrule
    \end{tabular}}
    \caption{F1-scores of AIA.}
    \label{aiaf1}
\end{table}

\begin{table}[!t]
    \centering
    \small
    \scalebox{0.90}{\begin{tabular}{c|cccc}
    \toprule
        Temprature & MRPC & RTE & AG\_News & YELP \\ \midrule
        0.0 & 0.7353 & 0.5668 & 0.8970 & 0.9485 \\ 
        0.5 & 0.7475 & 0.5415 & 0.9004 & 0.9499 \\ 
        1.0 & 0.7426 & 0.5343 & 0.9004 & 0.9503 \\ 
        5.0 & 0.7426 & 0.5343 & 0.8987 & 0.9512 \\ \bottomrule
    \end{tabular}}
    \caption{The success rate of model extraction attacks at different temperatures for the BERT model using the shadow data.}
    \label{temp1}
\end{table}

\begin{table}[!t]
    \centering
    \small
    \scalebox{0.90}{\begin{tabular}{c|cccc}
    \toprule
        Temprature & MRPC & RTE & AG\_News & YELP \\ \midrule
        0.0 & 0.7181 & 0.6029 & 0.8946 & 0.9466 \\ 
        0.5 & 0.7108 & 0.6029 & 0.8980 & 0.9473 \\ 
        1.0 & 0.7132 & 0.5848 & 0.8983 & 0.9464 \\ 
        5.0 & 0.7034 & 0.5487 & 0.8982 & 0.9478 \\ \bottomrule
    \end{tabular}}
    \caption{The success rate of model extraction attacks at different temperatures for the BERT model using the partial data).}
    \label{temp2}
\end{table}

\textbf{Model Extraction Attack (MEA)}: We show the performance of MEA in Figure~\ref{fig:mea} (taking temperature to be 0.5), and in general the attack achieves good results, for example, BERT model trained on YELP data achieves an accuracy of 0.950 (shadow data). Secondly, we can also see that the performance of partial data is generally lower than shadow data, one reason for this may be that the vector obtained by querying the target model with partial data has a lower entropy, which contains less information for the attacker to extract. We also report the impact of different temperatures on the attack, which can be seen in Table ~\ref{temp1} and ~\ref{temp2}. 

Tables ~\ref{temp1} and~\ref{temp2} show the success rate of the models in extracting attacks at different temperatures (BERT model), from which it can be seen that for models with suspected overfitting using the MRPC and RTE datasets, the experimental results tend to perform a little better when the temperature is lower. This suggests that they tend to prefer hard labels. In contrast, models with better generalization using the AG\_news and Yelp\_polarity datasets may tend to prefer labels with a higher temperature (softer). In general, the results of the experiments on different temperatures do not differ significantly.

\section{Performance using Auxiliary Data from Different Domains}
\label{16}
\textbf{Privacy Attacks}: In sub-figures 1-5 of Figure~\ref{fig:dd}, we show the attack performance of MIA, MDIA, and MEA for data from different domain scenarios. For both MIA and MEA attacks, we can obtain desirable results in most cases, even better than the case adopting the same data distribution hypothesis, such as in the black-box scenario when the target data is RTE and the data from different domains is MRPC (see the 1st subfigure). However, in some cases, the attack success rate is close to random guesses, e.g., for a black-box MIA attack, the attack success rate is 0.5 when the target data is MRPC and the data from different domains is SST2 (see the 1st subfigure). Another example is MEA, in which the target data is TREC and the data from different domains is SST2, the attack performance is 0.45 (see the 4th subfigure). For MDIA, the performance of attacks with the data from different domains is largely better than that of the first model inversion attack. This is also a reasonable observation since we obtain additional auxiliary data.

\begin{figure}[!h]
\centering
\subfigure{
\includegraphics[width=3cm]{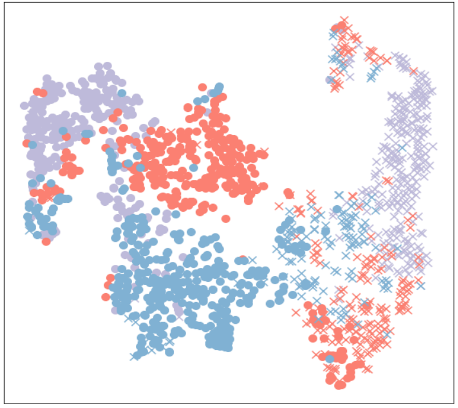}
}
\quad
\subfigure{
\includegraphics[width=3cm]{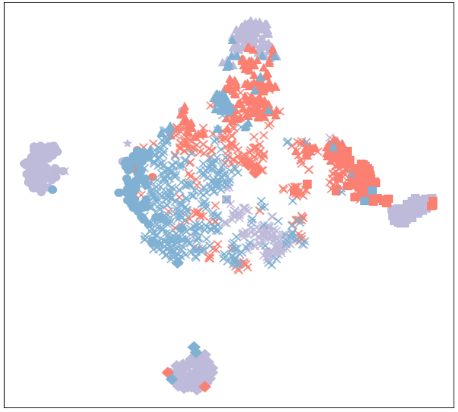}
}
\caption{\label{fig:dd_alysis}t-SNE plots for features obtained by querying the target model with data from different domains.}
\end{figure}

\begin{figure*}[!ht]
\centering
\subfigure{
\includegraphics[width=4cm]{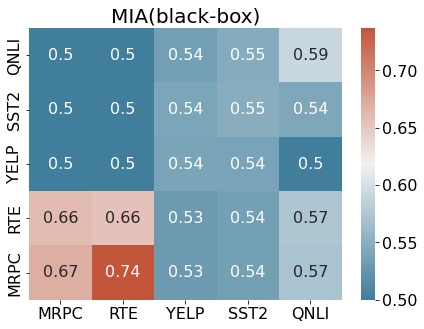}
}
\quad
\subfigure{
\includegraphics[width=4cm]{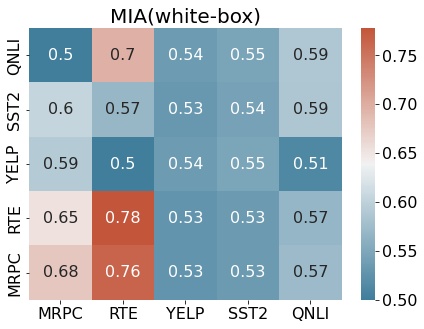}
}
\quad
\subfigure{
\includegraphics[width=4cm]{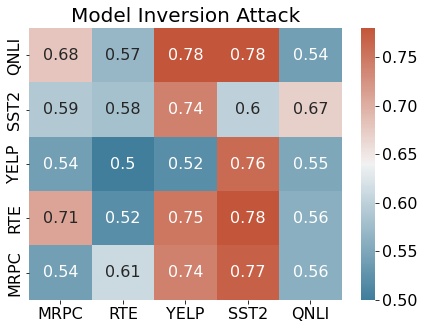}
}
\quad
\subfigure{
\includegraphics[width=4cm]{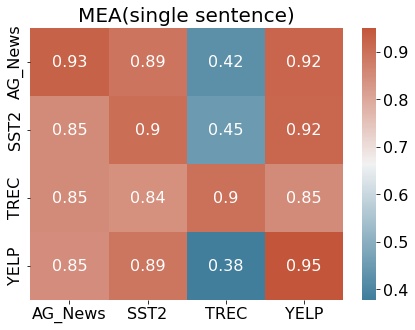}
}
\quad
\subfigure{
\includegraphics[width=4cm]{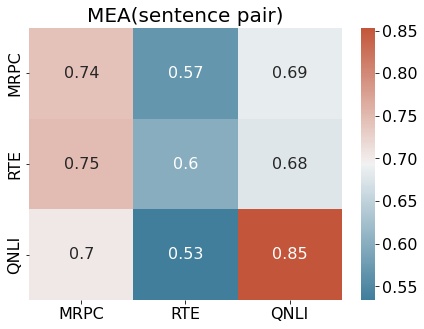}
}
\quad
\subfigure{
\includegraphics[width=4.2cm]{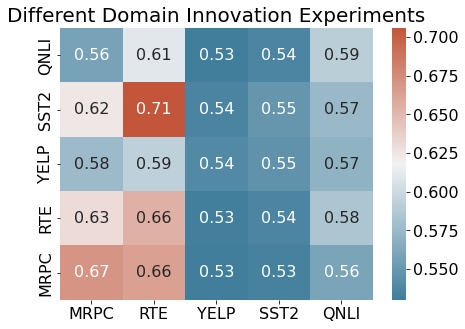}
}
\caption{\label{fig:dd}The experimental results show the performance of the BERT model on data from different domains. The horizontal axis represents the target training data, and the vertical axis represents data from different domains.}
\end{figure*}

\textbf{Cause Analysis}: To analyze the problem of lower attack performance with data from different domains, we plotted the t-SNE plots for features obtained by querying the target model with data from different domains, as shown in Figure~\ref{fig:dd_alysis}. In the first plot, the target data are SST2 (purple), and the data from different domains are AG\_News (red) and TREC (blue) respectively. The graph shows that the representation distribution is closer, so it will work better. However, it can be seen in the second graph that the distribution of representation of the data from different domains is far from the target data (the target data is TREC (purple) and the data from different domains is AG\_News (red) and SST2 (blue)), so this could explain why in some cases the attacker's success rate is lower.

\textbf{Experimental Results of Improved Methods}:
\label{erim}
Subfigure 6 in Figure~\ref{fig:dd} illustrates the results of experiments that mitigate the issue of the low success rate for membership inference under different domains. Compared with Subfigure 1, it can be seen that our proposed method of using the student model as a shadow model has great results. Among them, this method has significant improvement in the performance of membership inference attacks with a success rate close to 0.5. For example, when the target data is RTE and the data from different domains is QNLI, the success rate can reach 0.7.
\begin{table}[!t]
\small
  \centering
  \scalebox{0.90}{\begin{tabular}{c|cccc}
  \hline
  ~ & MRPC & RTE & AG\_News & YELP \\
  \hline
  Original & 0.7426 & 0.5812 & 0.9243 & 0.9498 \\
  DP-SGD($\epsilon$ = 5) & 0.6838 & 0.4729 & 0.8925 & 0.8987 \\
  DP-SGD($\epsilon$ = 15) & 0.7011 & 0.5050 & 0.8972 & 0.9040 \\
  SELENA & 0.7034 & 0.5740 & 0.9138 & 0.9533 \\
  TextHide & 0.7230 & 0.5126 & 0.9067 & 0.9214 \\
  \hline
  \end{tabular}}
  \caption{Performance of the target model using defense methods.}
  \label{tb2}
\end{table}

\begin{table}[!t]
    \centering
    \small
    \scalebox{0.90}{\begin{tabular}{c|ccc}
    \toprule
        ~ & AG(AIA) & BLOG & TP \\ \midrule
        Orignal & 0.793 & 0.939 & 0.865 \\ 
        DP-SGD($\epsilon$ = 5) & 0.741 & 0.882 & 0.834 \\ 
        DP-SGD($\epsilon$ = 15) & 0.745 & 0.891 & 0.840 \\ 
        SELENA & 0.785 & 0.918 & 0.866 \\ 
        TextHide & 0.764 & 0.906 & 0.795 \\ \bottomrule
    \end{tabular}}
    \caption{Performance of the target model using different defense methods.}
    \label{ta-po}
\end{table}

\begin{table*}[!ht]
    \centering
    \small
    \scalebox{0.80}{\begin{tabular}{c|ccccccccc}
    \toprule
        \multicolumn{2}{c}{} & \multicolumn{2}{c}{MRPC} & \multicolumn{2}{c}{RTE} & \multicolumn{2}{c}{AG\_News}  & \multicolumn{2}{c}{YELP} \\ \cline{3-10}
        \multicolumn{2}{c}{} & Original & $\epsilon$ = 5/15 & Original & $\epsilon$ = 5/15 & Original & $\epsilon$ = 5/15 & Original & $\epsilon$ = 5/15 \\ \midrule
        \multirow{4}{*}{MIA} & $<$Black,Shadow$>$ & 0.670 & 0.492/0.504 & 0.655 & 0.504/0.513 & 0.547 & 0.509/0.508 & 0.539 & 0.501/0.497 \\ 
        ~ & $<$Black,Partial$>$ & 0.662 & 0.493/0.509 & 0.829 & 0.513/0.548 & 0.557 & 0.506/0.509 & 0.54 & 0.500/0.501 \\ 
        ~ & $<$White,Shadow$>$ & 0.676 & 0.495/0.498 & 0.778 & 0.501/0.515 & 0.550 & 0.508/0.505 & 0.536 & 0.502/0.506 \\ 
        ~ & $<$White,Partial$>$ & 0.657 & 0.498/0.500 & 0.826 & 0.504/0.504 & 0.556 & 0.500/0.501 & 0.538 & 0.495/0.498 \\ \hline
        \multirow{2}{*}{MDIA} & $<$White,No$>$ & 0.56 & 0.54/0.50 & 0.54 & 0.50/0.50 & 0.28 & 0.26/0.26 & 0.54 & 0.50/0.52 \\ 
        ~ & $<$White,Shadow$>$ & - & - & - & - & 0.30/0.72 & 0.3/0.69 & 0.53/0.78 & 0.46/0.70 \\ \hline
        \multirow{2}{*}{MEA} & $<$Black,Shadow$>$ & 0.748 & 0.684/0.699 & 0.542 & 0.473/0.484 & 0.9 & 0.883/0.888 & 0.95 & 0.908/0.913 \\ 
        ~ & $<$Black,Partial$>$ & 0.711 & 0.683/0.699 & 0.603 & 0.473/477 & 0.898 & 0.881/0.884 & 0.947 & 0.910/0.913 \\ \bottomrule
    \end{tabular}}
    \caption{\label{tb3}Experimental results of defense methods against MIA, MDIA, and MEA using DP-SGD for the BERT model.}
\end{table*}
\begin{table*}[!ht]
    \centering
    \small
    \scalebox{0.80}{\begin{tabular}{c|ccccccccc}
    \toprule
        \multicolumn{2}{c}{} & \multicolumn{2}{c}{MRPC} & \multicolumn{2}{c}{RTE} & \multicolumn{2}{c}{AG\_News}  & \multicolumn{2}{c}{YELP} \\ \cline{3-10}
        \multicolumn{2}{c}{} & Original & Acc & Original & Acc & Original & Acc & Original & Acc \\ \midrule
        \multirow{4}{*}{MIA} & $<$Black,Shadow$>$ & 0.670 & 0.531 & 0.655 & 0.555 & 0.547 & 0.501 & 0.539 & 0.514 \\ 
        ~ & $<$Black,Partial$>$ & 0.662 & 0.500 & 0.829 & 0.608 & 0.557 & 0.498 & 0.540 & 0.511 \\ 
        ~ & $<$White,Shadow$>$ & 0.676 & 0.528 & 0.778 & 0.557 & 0.550 & 0.504 & 0.536 & 0.507 \\ 
        ~ & $<$White,Partial$>$ & 0.657 & 0.497 & 0.826 & 0.605 & 0.556 & 0.505 & 0.538 & 0.511 \\ \hline
        \multirow{2}{*}{MDIA} & $<$White,No$>$ & 0.56 & 0.58 & 0.54 & 0.50 & 0.28 & 0.26 & 0.54 & 0.52 \\ 
        ~ & $<$White,Shadow$>$ & - & - & - & - & 0.30/0.72 & 0.28/0.71 & 0.53/0.78 & 0.50/0.68 \\ \hline
       \multirow{2}{*}{MEA} & $<$Black,Shadow$>$ & 0.748 & 0.703 & 0.542 & 0.592 & 0.900 & 0.912 & 0.950 & 0.952 \\ 
        ~ & $<$Black,Partial$>$ & 0.711 & 0.704 & 0.603 & 0.545 & 0.898 & 0.909 & 0.947 & 0.951 \\ \bottomrule
    \end{tabular}}
    \caption{Experimental results of defense methods against MIA, MDIA, and MEA using SELENA for the BERT model.}
    \label{tb4}
\end{table*}
\begin{table*}[!ht]
    \centering
    \small
    \scalebox{0.80}{\begin{tabular}{c|ccccccccc}
    \toprule
        \multicolumn{2}{c}{} & \multicolumn{2}{c}{MRPC} & \multicolumn{2}{c}{RTE} & \multicolumn{2}{c}{AG\_News}  & \multicolumn{2}{c}{YELP} \\ \cline{3-10}
        \multicolumn{2}{c}{} & Original & Acc & Original & Acc & Original & Acc & Original & Acc \\ \midrule
        \multirow{4}{*}{MIA} & $<$Black,Shadow$>$ & 0.670 & 0.564 & 0.655 & 0.503 & 0.547 & 0.523 & 0.539 & 0.507 \\ 
        ~ & $<$Black,Partial$>$ & 0.662 & 0.570 & 0.829 & 0.510 & 0.557 & 0.527 & 0.540 & 0.506 \\ 
        ~ & $<$White,Shadow$>$ & 0.676 & 0.553 & 0.778 & 0.503 & 0.550 & 0.526 & 0.536 & 0.508 \\ 
        ~ & $<$White,Partial$>$ & 0.657 & 0.558 & 0.826 & 0.523 & 0.556 & 0.529 & 0.538 & 0.503 \\ \hline
        \multirow{2}{*}{MDIA} & $<$White,No$>$ & 0.56 & 0.56 & 0.54 & 0.50 & 0.28 & 0.28 & 0.54 & 0.50 \\ 
        ~ & $<$White,Shadow$>$ & - & - & - & - & 0.30/0.72 & 0.26/0.73 & 0.53/0.78 & 0.44/0.72 \\ \hline
       \multirow{2}{*}{MEA} & $<$Black,Shadow$>$ & 0.748 & 0.689 & 0.542 & 0.509 & 0.9 & 0.892 & 0.95 & 0.934 \\ 
        ~ & $<$Black,Partial$>$ & 0.711 & 0.683 & 0.603 & 0.524 & 0.898 & 0.893 & 0.947 & 0.936 \\ \bottomrule
    \end{tabular}}
    \caption{Experimental results of defense methods against MIA, MDIA, and MEA using TextHide for the BERT model.}
    \label{tb5}
\end{table*}
\begin{table*}[!ht]
    \centering
    \small
    \scalebox{0.80}{\begin{tabular}{c|ccccccc}
    \toprule
        \multicolumn{2}{c}{} & \multicolumn{2}{c}{AG\_News(AIA)} & \multicolumn{2}{c}{BLOG} & \multicolumn{2}{c}{TP}\\ \cline{3-8}
        \multicolumn{2}{c}{} & Original & Acc & Original & Acc & Original & Acc \\ \midrule
        \multirow{2}{*}{DP-SGD} & $<$Black,Shadow$>$ & 0.800 &    0.805/0.805 & 0.632 & 0.652/0.649 & 0.605 & 0.618/0.617 \\ 
        ~ & $<$White,Shadow$>$ & 0.947 & 0.945/0.947 & 0.659 & 0.649/0.694 & 0.681 & 0.617/0.682 \\ \hline
        \multirow{2}{*}{SELENA} & $<$Black,Shadow$>$ & 0.800 & 0.801 & 0.632 & 0.637 & 0.605 & 0.610 \\ 
        ~ & $<$White,Shadow$>$ & 0.947 & 0.947 & 0.659 & 0.673 & 0.681 & 0.690 \\ \hline
        \multirow{2}{*}{TextHide} & $<$Black,Shadow$>$ & 0.800 & 0.811 & 0.632 & 0.621 & 0.605 & 0.619 \\ 
        ~ & $<$White,Shadow$>$ & 0.947 & 0.923 & 0.659 & 0.639 & 0.681 & 0.649 \\ \bottomrule
    \end{tabular}}
    \caption{Experiment results of defense methods towards AIA for the BERT model. For DP-SGD, we report results with $\epsilon=5,15$.}
    \label{tb6}
    \vspace{-1.5em}
\end{table*}

\section{Defense Performance}
\label{17}
In this offensive and defensive system, we integrate three defensive methods, namely DP-SGD, SELENA, and TextHide. We report the effect on the BERT model. Table~\ref{tb2} shows the performance of the target model after it has been protected by defense methods (a table of attribute inference attack can be found in Table~\ref{ta-po}), where we can see that DP-SGD has a larger impact on performance.

\textbf{DP-SGD}: In Tables~\ref{tb3} and~\ref{tb6}, we report the effectiveness of DP-SGD's defense towards the four attacks. Overall, DP-SGD offers a more significant defense against the MIA, success rate of the attack is close to random guesses on the vast majority of the dataset. For example, on MRPC (black box/shadow data), DP-SGD has a defense capability of 0.492/0.504 against MIA, compared to an original attack performance is 0.670. For the other three attacks, the method is not significantly defensive but is still effective.

\textbf{SELENA}: Tables~\ref{tb4} and~\ref{tb6} show the effectiveness of SELENA's defenses, where the method is more effective against MIA, particularly MRPC and YELP. However, the defense against the other three attacks was poor, especially the AIA. For example, on the BLOG dataset, the original attack accuracy was 0.659, while SELENA performed 0.673. One possible reason for this is that the method does an aggregation operation on the output of multiple models, which allows the defense model to have more knowledge. Overall the method is not as good as DP-SGD at defending but has better identification accuracy than DP-SGD.

\textbf{Texthide}: In Tables~\ref{tb5} and~\ref{tb6}, we show the defensive capabilities of TextHide, which is effective mainly against MIA and MEA, and poorly against MDIA and AIA. For example, on RTE (black box/shadow data), the defense ability for the first two are 0.503 and 0.509 respectively, while the original attack performance is 0.655 and 0.542. As we can see, the defensive effectiveness of this method is comparable to SELENA, but of course, it is still better than SELENA's method on MDIA, AIA, and MEA. Overall the method has a good balance between accuracy and defensive ability.

\section{Results of the Chained Framework} 
\label{18}
Table ~\ref{tb7} describes the experimental results of the chained framework. It can be seen that the extraction model provides a strong defense against membership inference attacks, particularly for the YELP dataset, where the attack performance is nearly 0.5. This implies that if the model owner employs MEA and publishes the extracted model, it could serve as an effective defense method against MIA. When an attacker conducts an AIA on the extracted model in a white-box context, the attack performance is superior to that in a black-box scenario. Take the TP dataset as an example: the attack success rate can be improved by approximately 6 percentage points. This is primarily because, after the attacker gains access to the function-extracted model using a black-box method, they can employ white-box knowledge to intensify the attack.

From Table~\ref{tb7}, we can see that conducting a membership inference attack under the no-data condition is also feasible and yields a comparable attack performance to that achieved using same-distribution data.

We found that MIA enhances the performance of MEA on certain datasets (MRPC and RTE) when employed as a data filter. Furthermore, on most datasets, data generated by MDIA and filtered through MIA can effectively increase the success rate of MDIA. Both of these experimental outcomes are correlated with the success of MIA

\section{Experimental Results for LLM}
\label{19}
\textbf{ Membership Inference Attack for LLM (LMIA):} Table ~\ref{lmia} shows the AUC of membership inference attacks on LLM under different threat models. From the table, we can see that SPV-MIA (based on memorization) consistently outperforms all baseline methods across all LLM with different architectures and fine-tuning datasets. The underwhelming attack performance of LiRA and Neighbour Attack reveals their inability to be effectively applied to practical LLMs. This phenomenon verifies the claim that existing MIAs designed for LMs (based on overfitting) can not handle LLMs with large-scale parameters. Upon further analysis, we can also draw the following conclusion: The privacy risk caused by MIAs on LLMs is positively correlated with the overall NLP performance of the model itself.

\textbf{Model Inversion Attack for LLM (LMDIA):} Table ~\ref{tab:my_label} shows the number of memorized examples (out of 50 candidates) that we identify using each of the three text generation strategies and six membership inference techniques. In total across all strategies, we identify 315 unique memorized training examples from among the 900 possible candidates, for an aggregate true positive rate of 35\%. From the table, we can also see that the attacker's success rate using cross-domain data is higher than that of the no-data assumption. Out of privacy concerns, we do not display the recovered text in our paper, specific examples can be found on our GitHub.

\textbf{ Attribute Inference Attack for LLM  (LAIA):} Tables ~\ref{tab:attributes_accuracies}, ~\ref{tab:attributes} and ~\ref{tab:ce} present the results of LMIA under different threat models. From the tables, we can see that the capability of attribute inference attacks is correlated with the size of the model; the larger the model, the higher the success rate of the attribute inference attacks. Age and gender are attributes that are easier to infer, whereas education and occupation are more difficult to infer, likely because the latter represent more complex attribute information.

    \begin{table}[!t]
\centering
    \small
      \scalebox{0.90}{\begin{tabular}{c|cccc}
        \toprule
         \multirow{2}{*}{Model} & \multicolumn{2}{c}{CHIP-CTC} & \multicolumn{2}{c}{KUAKE-QIC}\\ \cmidrule{2-5}
         ~ &Original &ACC &Original &ACC\\
         \midrule
         Llama2 &0.793 & 0.782 &0.800 &0.788\\
         Qwen &0.796 & 0.772 &0.815 &0.806\\
      \bottomrule
    \end{tabular}}
    \caption{\label{llm-lmea-i-2}Attack Results of the LMEA-I ($D^{sha}$, P-Tuning v2).}
    \end{table}
    \begin{table}[!t]
    \centering
    \small
      \scalebox{0.90}{\begin{tabular}{c|cccc}
        \toprule
         \multirow{2}{*}{Model} & \multicolumn{2}{c}{CHIP-CTC} & \multicolumn{2}{c}{KUAKE-QIC}\\ \cmidrule{2-5}
         ~ &Original &ACC &Original &ACC\\
         \midrule
         Llama2 &0.793 & 0.788 &0.800 &0.801\\
         Qwen &0.796 & 0.785 &0.815 &0.820\\
      \bottomrule
    \end{tabular}}
    \caption{\label{llm-lmea-i-4}Attack Results of the LMEA-I ($D^{par}$, P-Tuning v2).}
    \end{table}

\textbf{ Model Extraction Attack for LLM (LMEA):} 
In Table~\ref{llm-lmea-g}, we present the experimental results for LMEA-G. The data from Table~\ref{llm-lmea-g} indicate that Qwen is more vulnerable to the LMEA-G attack, consequently revealing more of its knowledge. In Tables~\ref{llm-lmea-i}, ~\ref{llm-lmea-i-2}, ~\ref{llm-lmea-i-3}, and ~\ref{llm-lmea-i-4}, we display the experimental results for LMEA-I. We can observe that, regardless of the threat model and fine-tuning strategy employed, attackers can easily transfer knowledge from the target model to the extraction model. Based on the comparison of the two types of LMEA mentioned above, we can infer that general LLMs possess a stronger capacity to resist LMEA compared to domain-specific LLMs. 





\end{document}